\documentclass{cta-author}

{}
{}
{}

\begin{document}

\supertitle{Submission Template for IET Intelligent Transport Systems}

\title{Vision-based Vehicle Speed Estimation: A Survey}

\author{\au{David Fern\'andez Llorca$^{1\corr,2}$}, \au{Antonio Hern\'andez Mart\'inez$^{1}$}, \au{Iv\'an Garc\'ia Daza$^{1}$}}

\address{\add{1}{Computer Engineering Department, University of Alcal\'a, University Campus, Alcal\'a de Henares, 28805, Madrid, Spain}
\add{2}{European Commission, Joint Research Center, Seville, Spain}
\email{david.fernandez-llorca@ec.europa.eu}}

\begin{abstract}
The need to accurately estimate the speed of road vehicles is becoming increasingly important for at least two main reasons. First, the number of speed cameras installed worldwide has been growing in recent years, as the introduction and enforcement of appropriate speed limits are considered one of the most effective means to increase the road safety. Second, traffic monitoring and forecasting in road networks plays a fundamental role to enhance traffic, emissions and energy consumption in smart cities,  being the speed of the vehicles one of the most relevant parameters of the traffic state. Among the technologies available for the accurate detection of vehicle speed, the use of vision-based systems brings great challenges to be solved, but also great potential advantages, such as the drastic reduction of costs due to the absence of expensive range sensors, and the possibility of identifying vehicles accurately. This paper provides a review of vision-based vehicle speed estimation. We describe the terminology, the application domains, and propose a complete taxonomy of a large selection of works that categorizes all stages involved. An overview of performance evaluation metrics and available datasets is provided. Finally, we discuss current limitations and future directions. 
\end{abstract}

\maketitle

\section{Introduction}\label{sec1}

The ability to provide an accurate estimate of the speed of road vehicles is a key feature for Intelligent Transport Systems (ITS). This requires tackling problems such as synchronized data recording, representation, detection and tracking, distance and speed estimation, etc., and it can be addressed using data from sensors of different nature (e.g., radar, laser or cameras) and under different lighting and weather conditions. 

This is a well-known topic in the field with considerable contributions from both academia and industry. The most important application domains are speed limit enforcement, traffic monitoring and control, and, more recently, autonomous vehicles. However, it is important to note that the nature of the problem of speed detection differs in a not insignificant way, depending on whether it is done from the infrastructure (fixed sensors) or from a mobile platform (intelligent/autonomous vehicles or mobile speed cameras). Our study focuses on fixed systems from the infrastructure, although many of the presented methodologies can be easily applicable to the case of mobile sensors by compensating the speed of the ego-vehicle.

The requirements concerning the uncertainty and robustness on the vehicle speed estimation from the infrastructure depends on the application scenario. On the one hand, for speed enforcement applications, extreme maximum permissible error standards are required by national or international certification agencies [1], which is more than reasonable considering that these automatic systems are used to fine, or even imprison, drivers who exceed the maximum speed limit. This implies that the most common sensors used for speed measurement in these applications are based on high-precision high-cost range sensors, such as radar (Doppler effect) and laser (time of flight), or, in some cases, accurate on-pavement sensors (e.g., inductive or piezoelectric loop detectors) embedded in pairs under the road surface. The use of visual cues to measure physical variables related with the motion of the vehicles has been rarely proposed. Video cameras usually play a secondary role as a means of capturing a human-understandable representation of the scene, allowing the visual identification of the vehicles, and serving as evidence for speed limit enforcement. On the other hand, for traffic monitoring, forecasting and control applications, there is no regulatory framework or scientific consensus on the accuracy and uncertainty of speed estimation [2]. The requirements regarding sensors are therefore reduced, and it is more common to see the use of less accurate sensors [3] including traffic cameras [4]. In any case it is obvious that the higher the accuracy of the speed estimation, the better the forecasting and control approaches will be. 

Recent advances in computer vision techniques have led to a significant increase in the number of works proposing the use of vision as the only mechanism for measuring vehicle speed. The challenge of making accurate estimations of distances and speeds arises from the discrete nature of video sensors that project the 3D world into a 2D discrete plane. This intrinsic limitation results in a digital representation whose accuracy follows an inverse-square law, i.e., its quantity is inversely proportional to the square of the distance from the camera to the vehicle. Despite these limitations, the potential benefits of using video cameras are remarkable, since they are a cost-efficient solution compared with range sensors such as microwave Doppler radars or laser-based devices. The possibility of using already installed traffic cameras without the need to integrate new sensors is another advantage. Also, the visual cues from these sensors can be used to address other problems such as automatic identification of the vehicle (license plate, make, model and color), re-identification in different places, and if the resolution allows, the identification of occupants, or seat belt use. 


The scope of this survey is vision-based vehicle speed detection from the infrastructure, including fixed point speed cameras and traffic cameras, and excluding mobile speed cameras or onboard cameras for intelligent vehicle applications. Despite the potential impact, and the high number of contributions available, there is still no detailed survey of this topic. We describe the terminology and the application domains. We survey a large selection of works and propose a complete and novel taxonomy that categorizes all the components involved in the speed detection process. Performance evaluation metrics and available datasets are examined. Finally, current limitations are discussed and major open challenges for future research are outlined. We hope that this work will help researchers, practitioners and policy makers to have a detailed overview of the subject that will be useful for future work.

The structure of this paper is as follows: we present the terminology and describe the application domains in Sec. \ref{sec:overview}, we present a general description of the taxonomy in Sec. \ref{sec:tax}, and we review and analyze each of its components in Secs. \ref{sec:t1}-\ref{sec:data-metrics}. The description of the available datasets and evaluation metrics are provided in Sec. \ref{sec:data-metrics}. Discussion of the current state of the art and future trends are provided in Sec. \ref{sec:discussion}, and, finally, Sec. \ref{sec:conclusions} concludes the presented work.

\section{Overview and terminology}\label{sec:overview}
The main components of a fixed vision-based vehicle speed measurement system are summarized in Fig. \ref{fig:1}. Four levels of abstraction are considered, which are linked to the main terminology in what follows:

\begin{figure}[!t]
	\centering	
	\includegraphics[width=0.48\textwidth]{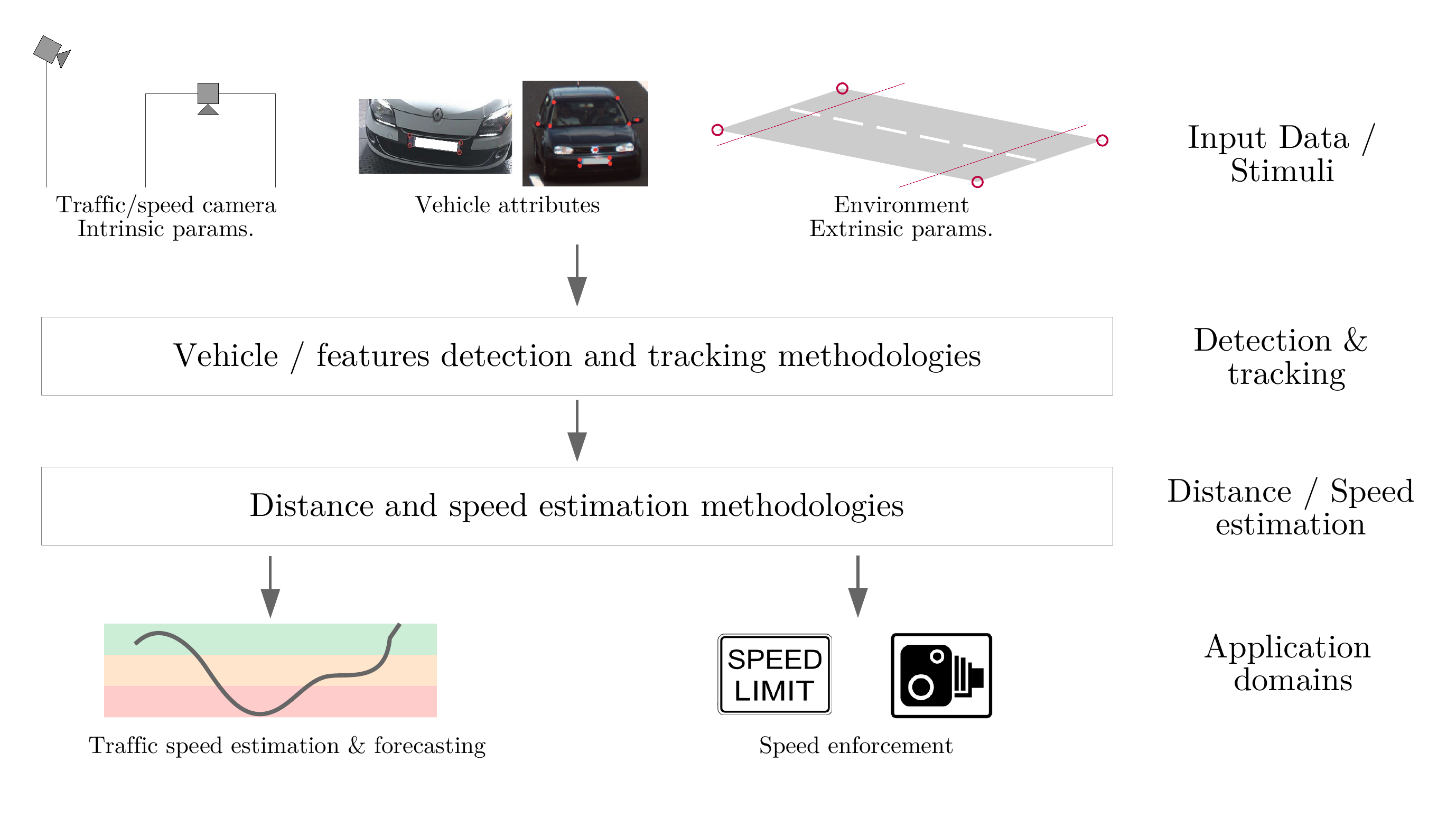}
	\caption{Main high-level components of a vision-based vehicle speed estimation system: input data, object detection and tracking, distance and speed estimation, and outcome.}
	\label{fig:1}
\end{figure}

\subsection{Input data / stimuli}
Vision-based speed detection approaches are based on input data from video cameras. For each vehicle there will be a sequence of images from the first appearance to the last one. The number of images available will depend on the camera pose with respect to the road, the focal length, the frame rate, and the speed of the vehicle. Two different types of video cameras can be considered: 

\begin{itemize}
\item{\emph{Traffic cameras}}: also referred as traffic surveillance cameras or traffic CCTV. They are used to monitor traffic flow, congestion and accidents both manually (visual inspection by a human operator) or automatically. In most cases, these are fixed cameras located far from the traffic flow, infrastructure- or drone-based. 

\item{\emph{Speed cameras}}: also referred as traffic enforcement camera. They are generally considered to be a device for monitoring the speed of vehicles independently of the sensor used to measure the speed (radar, laser or vision). Even for radar- or laser-based systems, there is always a camera to record an image of the vehicle, and hence the term \emph{camera}. In this case, we use the term speed camera in the most literal way, i.e. vision-based systems for measuring speed. Their location is usually closer to the traffic flow than the traffic cameras.  
\end{itemize}

Other forms of input data include vehicle attributes such as vehicle type, keypoints, license plate size, etc. Camera calibration plays a key role in providing both intrinsic and extrinsic parameters. Prior knowledge of the size of the road section reveals information that is critical to calculating the extrinsic relationship between the road and the camera or even the speed of the vehicles. 

\subsection{Detection and tracking} 
The vehicles, or some of their representative features, must be detected in all the available images. Tracking the vehicle or its features over time is essential to obtain speed measurements. Different approaches can be applied to address both tasks. 

\subsection{Distance and speed estimation} 
Speed estimation intrinsically involves the estimation of distances with associated timestamps. Different methods exist to calculate the relative distance of vehicles from some global reference, as well as different approaches to finally calculate the speed of the vehicle.   

\subsection{Application domains}
\begin{figure}[!t]
	\centering	
	\includegraphics[width=0.48\textwidth]{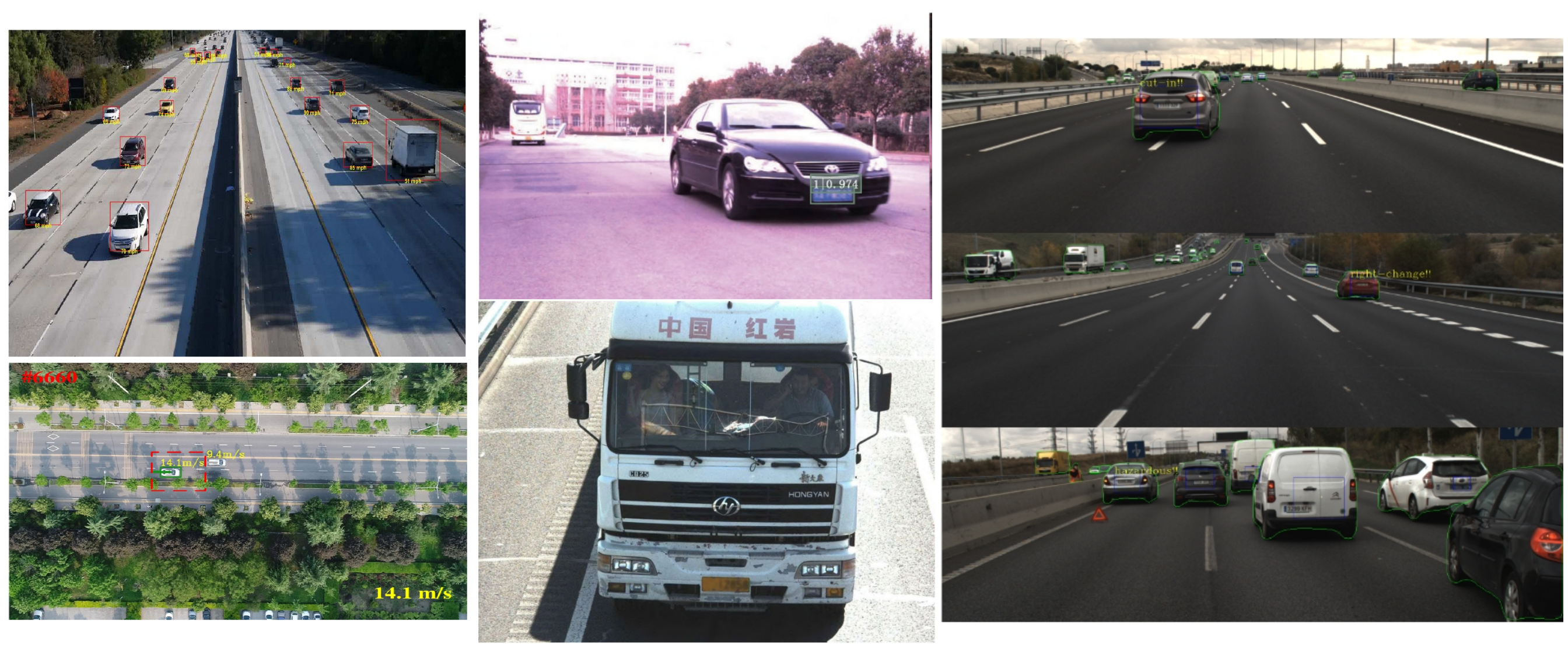}
	\caption{Examples of images from different application domains. Left column: traffic and drone cameras. Middle column: speed cameras. Right column: autonomous vehicles context. Images obtained from [88], [100], [105] and [130].}
	\label{fig:2}
\end{figure}

Vision-based vehicle speed estimation is a fundamental task for traffic monitoring, forecasting and control applications, speed enforcement, and, in addition, autonomous or intelligent vehicles (different examples are depicted in Fig. \ref{fig:2}). 

\subsubsection{Traffic monitoring and control}
The knowledge of the speed of all the vehicles in multiple road segments allows to know, and even forecast, the traffic state of the road (flow, density and speed), which can be used for a very wide set of different applications such as adaptive traffic signal control, real-time traffic aware navigation, traffic jams, congestion and accident detection, as well as other off-line operation and planning scenarios such as infrastructure enhancement.  

\subsubsection{Speed enforcement}
Speed has been identified as a key risk factor in road traffic injuries, increasing both the risk of road traffic crashes and the severity of the consequences [5]. And speed enforcement is a key road safety measure that leads to a reduction in accidents [6]. As suggested in [7], in the vicinity of speed cameras the reduction of speeding vehicles and crashes can reach up to $35\%$ and $25\%$ respectively. Moreover, the higher the intensity or level of enforcement, the larger the accident reduction [8].

Depending on the country, speed offenders can be fined, lose their driver's license, or even go to jail.  Therefore, speed estimation technology must be extremely accurate. National and international metrology or measurement authorities impose high standards of accuracy that must be met to allow the use of speed cameras to enforce speed limits. The maximum permissible errors (MPE) are usually defined as: 

\begin{equation}
    MPE \leq
    \begin{cases}
      \pm N, & \text{if}\ S \leq 100km/h \\
      < N\%, & \text{if}\ S > 100km/h
    \end{cases}
    \label{eq:req}
 \end{equation}
where $S$ is the actual speed of the vehicle, and $N$ is a value usually defined between 1 and 3 depending on the country. Designing vision-based vehicle speed estimation systems that meet these accuracy requirements is certainly a complex challenge due to the intrinsic limitations of vision sensors for measuring distances. 

\subsubsection{Autonomous vehicles} 
The ability to estimate the relative distances and speeds of surrounding vehicles is essential for advanced driver assistance systems and automated driving. From simple Adaptive Cruise Control (ACC) systems [9], through Traffic Jam Assist and Chauffeur systems that combine ACC Stop \& Go [10] and Lane Keeping Assist, to sophisticated Highway Chauffeur or Autopilot system which include complex maneuvers such as overtaking [11], the accurate localization of target vehicles and the estimation of their relative speed are fundamental variables used by low-level control systems. Vision-based approaches have been widely addressed to deal with vehicle detection, using flat-world assumption, stereo-vision or by combining cameras with range-based sensors such as radar or LiDAR. Although vehicle speed estimation for autonomous vehicles is out of the scope of this survey, most of the analyzed approaches can be extended to be used on-board of mobile platforms.  

\section{Taxonomy}\label{sec:tax}
\begin{figure*}[!t]
	\centering	
	\includegraphics[width=0.95\textwidth]{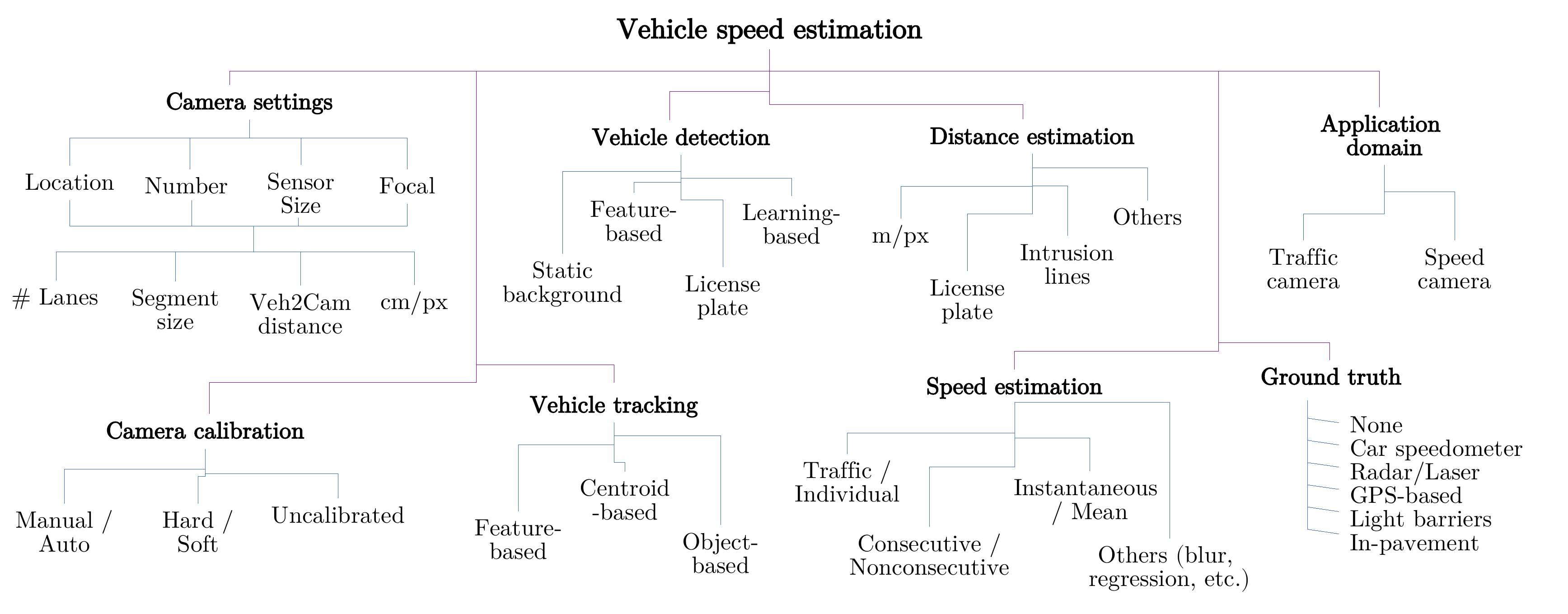}
	\caption{Proposed vision-based vehicle speed estimation taxonomy.}
	\label{fig:taxonomy}
\end{figure*}

In this survey, more than 135 papers related with vision-based vehicle speed detection have been evaluated. To provide a consistent numbering of bibliographic references, these are provided in temporary order by year. The proposed taxonomy breaks down the problem of speed detection, including all the components, approaches and experimental details involved (see Fig. \ref{fig:taxonomy} for a general overview). We start with the input stimuli and explain how camera settings and calibration procedures affect the subsequent steps. Next, we analyse all the steps involved, from vehicle detection and tracking to distance and speed estimation. Finally, we discuss the different approaches to evaluate the accuracy of the speed measurements.

On the one hand, the taxonomy presented is intended to serve as a basis for understanding what the main elements and issues are from a high-level perspective (e.g. practitioners and policy makers). On the other hand, it is intended to serve as a tool for the development of future approaches for future research.

\subsection{Application domain}
The first categorization is directly related to the attempt to devise an accurate system to be used for speed enforcement (\textbf{speed camera} application domain) [13, 17, 18, 32, 33, 39, 41-43, 51, 53, 54, 61, 65-68,  70, 72, 74, 76-78, 80, 82, 84-86, 94, 100-102, 107, 108, 112, 114, 119, 122-125] or a less accurate approach that can still be used for traffic speed estimation (\textbf{traffic camera} application scenario) [12, 14-16, 19-31, 34-38, 40, 44-50, 52, 55-60, 62-64, 69, 71, 73, 75, 79, 81, 83, 87-93, 95-99, 103-106, 109-111, 113, 115-118, 120, 121, 126-129]. Although in some cases it is claimed that accurate speed detection is targeted, some aspects such as the low resolution of the camera, a limited experimental evaluation or the absence of an appropriate ground truth make us to categorize these manuscripts as traffic speed systems. 

\begin{figure}[!t]
	\centering	
	\includegraphics[width=0.48\textwidth]{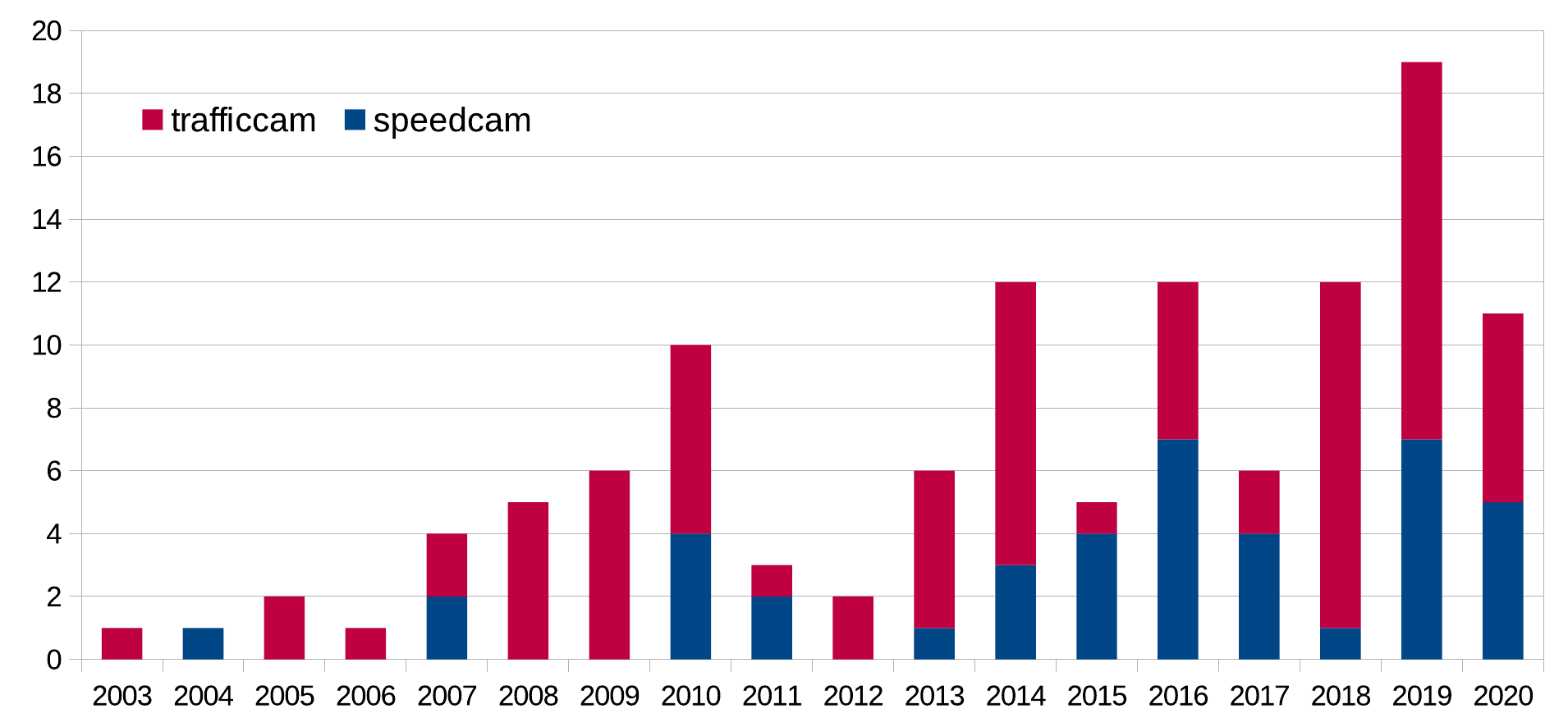}
	\caption{Publication trends in the state of the art by application domain.}
	\label{fig:years}
\end{figure}

With this categorization, in Fig. \ref{fig:years} we present the distribution of the papers published by year. As can be seen, the general trend in the number of publications is upwards, with the number of proposals focusing on accurate speed detection for speed enforcement being much lower than the number of proposals focusing on traffic speed detection.

\subsection{Camera settings}
This category involves the camera intrinsic (sensor size and resolution, focal length) and extrinsic (location with respect to the road plane, including drone-based cameras) parameters, and the number of cameras (mono, stereo or multiple). Depending on these parameters, the real-world scenario projected on the image plane can represent one or several lanes along a short or large road segment. Depending on this configuration, and the relative position of the vehicles with respect to the camera, we finally have one of the most important variables: the \textbf{meter-to-pixel ratio}, i.e., the portion of the road segment represented by each pixel. The lower this ratio is, the greater the accuracy in estimating distance and speed. Due to the perspective projection model, this ratio is proportional to the square of the distance from the camera, which means that those measurements taken at long distances have very poor accuracy. Note that we omit integration details such as the exact location of the camera, anchoring systems, type of gantry, pole or bridge, etc. These features are rarely provided in the reviewed works.

\subsection{Camera calibration}\label{sec:camcal}
An accurate estimation of the camera intrinsic and extrinsic parameters is required to provide measurements in real-world coordinates. The most common approach is to perform a \textbf{soft calibration}, i. e., intrinsic parameters are calibrated in the lab or approximated using the sensor and lens characteristics, and extrinsic parameters (rigid transformation between the road plane and the camera) are obtained using a manual [17, 28, 31, 35, 37, 47, 54, 59, 66, 67, 70, 74, 76, 80, 83, 86, 91, 92, 94, 100, 109, 110, 117, 119, 121, 125, 128, 129] or an automatic procedure [77, 90]. \textbf{Hard calibration} involves to jointly estimate both intrinsic and extrinsic parameters with the camera already installed. It can also be performed either manually [20, 38, 58] or automatically [12, 14, 18, 19, ,23, 33, 78, 85, 88, 101, 108]. In some limited cases, camera calibration is somehow neglected [13, 15, 16, 21, 22, 24-26, 30, 32, 34, 39-41, 43, 44, 46, 48-53, 55, 56, 60-65, 69, 71, 73, 81, 84, 95, 103-107, 111, 113, 118, 120, 124]. 

\subsection{Vehicle detection}
Since the cameras are mostly static (except for the drone cameras), in most cases, vehicle detection is addressed by modeling and subtracting the background [12-15, 18, 20-28, 30, 32, 34-37, 40, 42-44, 47, 48, 50, 51, 53-55, 61-63, 65, 71-75, 81, 83, 84, 86, 92, 93, 95, 102, 103, 108, 110-113, 117 118, 121]. Other approaches are feature-based, for example, detecting the license plate [31, 41, 49, 60, 66, 68, 70, 76, 77, 80, 82, 94, 122, 125] or other features of the vehicle [38, 39, 45, 49, 57-59, 69, 71, 79, 87, 91, 95, 96, 98, 99, 110-112, 116, 117, 123]. The use of learning-based methods to recognize vehicles in images has been growing recently [19, 85, 88-91, 98, 100, 101, 104-106, 115, 120, 126-129].

\subsection{Vehicle tracking}
The ability to have a smooth and stable trajectory of the vehicles is a key issue to deal with vehicle speed detection. We can classify vehicle tracking into three different categories. First, feature-based [17, 19, 20, 22, 31, 33, 37, 38, 45, 51, 53, 55, 58, 59, 61, 62, 69, 71, 74, 79, 85, 88, 91, 92, 93, 97, 112], which tracks some set of features from the vehicle (e.g., optical flow). Second, those methods focused on tracking the centroid of the blob or the bounding box of the vehicle [25, 34, 35, 40, 48, 50, 52, 56, 65, 72, 73, 75, 83, 86, 95, 103, 106, 111, 118, 121, 122]. Finally, those approaches that focus on tracking the entire vehicle [15, 16, 18, 26-30, 32, 36, 42-44, 47, 54, 57, 63, 78, 81, 84, 89, 91, 98, 102, 104, 105, 108, 109, 111, 113, 114, 120, 125, 128, 129], or a specific part of it (such as the license plate [41, 49, 60, 66, 68, 70, 76, 77, 80, 82, 94, 100, 125]). 

\subsection{Distance estimation}
For monocular systems, the estimation of the distance of the vehicles is usually computed with a set of constraints such as flat road assumption, including methods based on the homography [15, 16, 18, 20, 22, 34, 37, 52, 53, 61, 66, 94, 98, 99, 110] and the use of augmented intrusion lines, patterns or regions [19, 20, 23, 24, 26, 32, 38, 47, 79, 86, 87, 107, 124], or by using prior knowledge about the actual dimensions of some of the objects (e.g., the license plates [41, 49, 60, 66, 68, 80, 125] or the size of the vehicles [42, 62, 78, 103]). These restrictions are alleviated when using stereo vision [42, 60, 61, 74, 80, 92, 100, 112].  



\subsection{Speed estimation}
In a few cases, the problem of detecting the speed of vehicles is posed as a problem of detecting the speed of traffic on a segment of the road, i.e. directly obtaining an average road speed value [12, 17, 20, 23, 32, 34, 38, 40, 42, 51, 53, 54, 58, 61, 69, 77, 79, 80, 81, 84, 85, 94, 102, 105, 110, 113, 121]. However, in most cases the detection is carried out at the level of individual vehicles. Prior knowledge about the frame rate of the camera or accurate timestamps per each image are needed to compute the time between measurements. The use of consecutive [14, 18, 19, 22, 25, 33, 35-37, 41, 46, 48, 50, 55, 65, 67, 68, 71-76, 78, 82, 83, 85, 88, 89, 92, 96, 100, 103, 120, 129] or non-consecutive [15, 24, 26, 28, 30, 31, 43, 44, 47, 49, 52, 59, 66, 70, 71, 86, 87, 93, 95, 99, 106-108, 111, 118, 124, 125, 128] images to estimate speed is a fundamental variable that has a considerable impact on accuracy. How to integrate all available measurements (instantaneous, mean, optimal, etc.) is also a key factor affecting the final accuracy of the measurement. 

\subsection{Ground truth generation}
\label{subec:gt}
To evaluate the accuracy of the different vehicle speed estimation methodologies, an appropriate ground truth of the actual vehicle speed is required. Surprisingly, a considerable number of works presented their results without comparing them with reference obtained with a higher precision system. This raises questions about the validity of the results presented. 

As can be observed in Fig. \ref{fig:taxonomy}, the technologies used to compute the ground truth include the following: car speedometers, range-based technologies such as radar or laser, GPS-based including standard GPS, GPS odometers, and DGPS, light barriers and in-pavement sensors such as piezoelectric or inductive. 

Even though it is more of a technological problem than a scientific one, the synchronization of the data from the cameras and the other sensors is a critical and complex issue [80]. The sensors are usually located in different places, with independent recording systems, and with different sampling frequencies.

\section{Camera settings and calibration}\label{sec:t1}

\subsection{Camera settings}
The camera settings have a direct impact on the accuracy of the speed estimation method. The first categorization considered is the camera position, including drone cameras [25, 30, 57, 62, 64, 69, 96, 104-106, 126], and traffic cameras that can be classified roughly according to their height as distant ($\geq 5m$) or close ($<5m$) traffic cameras. In some cases the cameras are located on the side of the road, providing side views of the vehicles [13, 21, 26, 33, 39, 73, 112, 118]. Most works are based on monocular systems, but we can also find a few approaches based on stereo [42, 60, 74, 100, 112] and non-stereo multi-camera [61, 80] setups. Although most approaches are based on conventional cameras with CCD or CMOS sensors, we can find some works using event-based cameras [16, 17], based on bio-inspired sensors [147].

Considering camera resolution one can expect to have a progressive increase of the pixel resolution over time, mainly due to advances in hardware. However, this is not the case, and a considerable number of modern works (since 2010 we have [32-35, 42-45, 52, 54, 67, 71-73, 79, 83, 87, 97, 103, 117]) with a resolution lower than or equal to 640$\times$480 pixels (VGA). Resolution not only affects the accuracy of image processing detection techniques, but also the accuracy of distance estimation. The higher the resolution, the lower the meter-to-pixel ratio. 

The focal length is also a fundamental parameter that is usually linked to the height of the camera, the length of the road segment and the number of lanes covered by the field of view. As explained in Sections \ref{sec:disterror} and \ref{sec:speederror}, for a specific camera height, the larger the focal length the higher the accuracy of the distance and speed measurements. 

Most works use a low-medium focal length ($\leq 25$mm) since they are designed to cover multiple lanes and a large road stretch. Only a few works implicitly or explicitly highlight the need for a long focal length to increase the accuracy of the speed estimation, even if only one lane is covered by the system [23, 31, 53, 61, 68, 70, 76, 77, 80, 82, 107, 124]. 

\begin{figure}[!t]
	\centering	
	\includegraphics[width=0.48\textwidth]{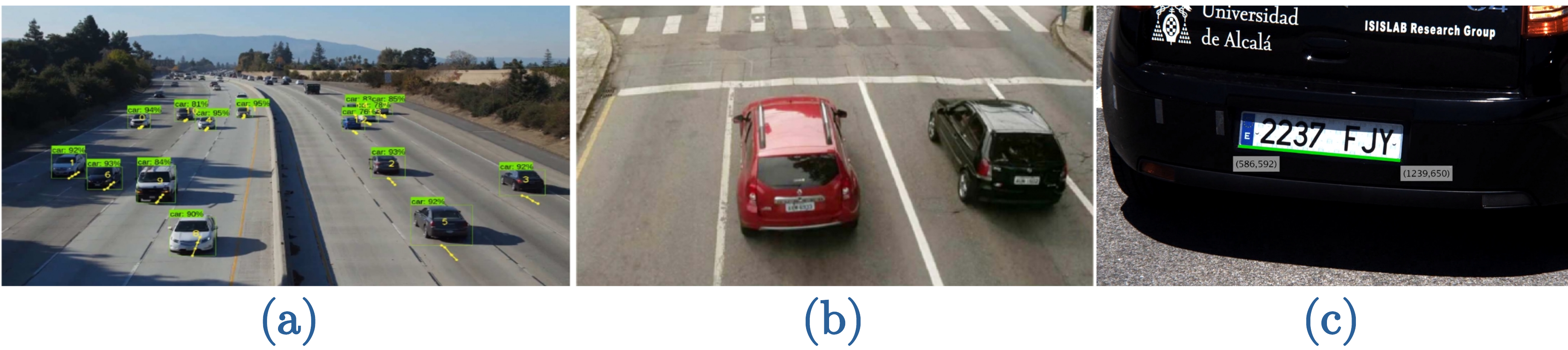}
	\caption{Examples of camera settings. (a) Low-medium focal length, very far relative distances. Multiple lanes, both directions. Very high meter-to-pixel ratio [88]. (b) Medium focal length, medium relative distances. Multiple lanes, one direction. Medium meter-to-pixel ratio [82]. (c) High focal length, low relative distances. One lane, one direction. Low meter-to-pixel ratio [80].}
	\label{fig:cameras}
\end{figure}

Therefore, depending on the camera intrinsics and extrinsics (location w.r.t. the road) parameters we finally have different scenarios projected into the image plane. As depicted in Fig. \ref{fig:cameras}, we can roughly define three type of scenarios:

\begin{itemize}
\item \emph{High meter-to-pixel ratio} (Fig. \ref{fig:cameras}.a): obtained with low-medium camera resolution, low focal lengths, high camera height, covering a long road segment, multiple lanes and both directions. Some examples are [12, 15, 19, 22, 27, 36, 48, 74, 75, 81, 88, 121, 128, 129].
\item \emph{Medium meter-to-pixel ratio} (Fig. \ref{fig:cameras}.b): obtained with medium-high camera resolution, medium focal lengths, medium camera height, covering a medium road segment, multiple lanes and only one direction. Some examples are [18, 41, 42, 45, 54, 61, 76, 82, 93, 108].
\item \emph{Low meter-to-pixel ratio} (Fig. \ref{fig:cameras}.c): obtained with high camera resolution, high focal lengths, low camera height, covering a short road segment, one or two lanes and only one direction. Some examples are [23, 31, 53, 68, 70, 77, 80, 107, 124]. 
\end{itemize}

\subsection{System calibration}
\label{subsec:cal}
Accurate camera calibration is of utmost importance for vision-based distance and speed estimation of objects. Whether assuming that vehicles are moving along a flat road for monocular systems or using a stereo configuration, the ability to obtain accurate real-world measurements from pixel coordinates considerably depends on accurate system calibration. For monocular systems, the pinhole camera model provides a general expression to describe the geometric relationship between a 2D pixel in the image plane $(u,v)$ and a 3D real-world coordinate system $(X_w, Y_w, Z_w)$, in this case, placed on the road plane and assuming a flat road shape:

\begin{equation}
\left[ \begin{array}{c} su \\ sv \\ s \end{array} \right] = 
\left[ \begin{array}{ccc} f_x & 0 & c_x \\ 0 & f_y & c_y \\ 0 & 0 & 1 \end{array} \right]
\left[ \begin{array}{cccc} r_{11} & r_{12} & r_{13} & t_x \\ r_{21} & r_{22} & r_{23} & t_y \\ r_{31} & r_{32} & r_{33} & t_z \end{array} \right]
\left[ \begin{array}{c} X_w \\ Y_w \\ Z_w \\ 1 \end{array} \right]
\end{equation}
where $f_x$ and $f_y$ are the focal distances along the $x$ and $y$ axes of the camera respectively, $c_x$ and $c_y$ represent the image optical center. These are the intrinsic parameters that can be represented in the intrinsic matrix $K$, which assumes that the distortion of the image has been appropriately corrected (note that the use of large focal lengths considerably reduces the effect of the distortion). The rotation and translation parameters can be grouped in a $3\times 3$ rotation matrix $R$ and a $3\times 1$ translation vector $T$. 

\begin{figure}[!t]
	\centering	
	\includegraphics[width=0.48\textwidth]{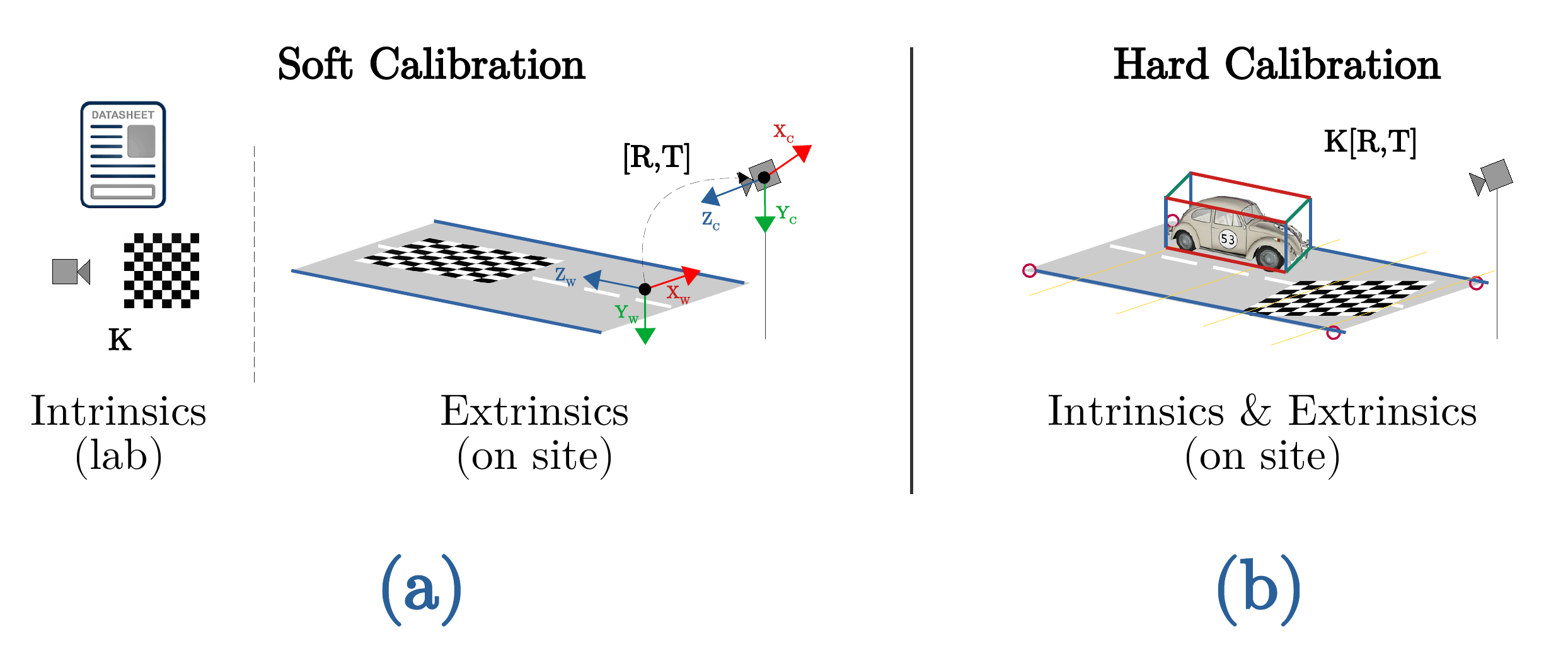}
	\caption{Camera calibration procedures. (a) \textbf{Soft calibration}. Intrinsic parameters (focal distance, optical center and distortion coefficients) are computed in the lab. Then, for a fixed focal length, the extrinsic parameters are computed on-site. (b) \textbf{Hard calibration}. Both intrinsic and extrinsic parameters are computed on site. In both cases the procedures can be manual or automatic. }
	\label{fig:calhs}
\end{figure}

The monocular camera calibration problem is the problem of estimating $K$, $R$ and $T$. As stated in Section \ref{sec:camcal}, and depicted in Fig. \ref{fig:calhs}, two main approaches are applied when dealing with camera calibration for vehicle speed estimation. First, \textbf{soft} calibration (Fig. \ref{fig:calhs}.a), which estimates $K$ in the lab using calibration patterns (e.g., chessboard) or assumes the parameters provided in the datasheets (sensor resolution, pixel size and focal length), and then computes $[R,T]$ on site. Second, \textbf{hard} calibration (Fig. \ref{fig:calhs}.b) which calculates $K$ and $[R,T]$ on site. 

One of the most common ways to compute the camera's extrinsic parameters in the operation environment is using vanishing points [12, 14, 15, 19, 22, 23, 27, 33, 42, 85, 88, 89, 91, 108]. The set of parallel lines in the 3D real-world coordinate system intersect at a unique 2D point when projected to the image plane. This point is commonly referred to as the vanishing point. The obvious set of lines in this case are the road  markings (including solid and dashed lane lines) and the road edges that are usually of high contrast. The location of the vanishing point allows the computation of the rotation matrix $R$. The translation matrix $T$ is then obtained using knowledge of real-world dimensions of some object or region in the image. Another common approach to perform calibration of extrinsic parameters is using known geometric static features present on the road plane and then compute the homography transformation which includes rotation, translation and scale [31, 38, 47, 66, 67, 70, 72, 75, 79, 82, 86, 87, 98, 99, 110, 113, 119, 121, 128, 129]. The dimensions, distances, lengths, etc., of these static features have to be known in advance. This may involve the need to stop traffic to access the operating environment and take direct measurements or use some indirect method (e.g., using laser scanners, or even Google Maps).

Hard calibration to get both camera's intrinsic and extrinsic parameters can be addressed by using standard (e.g., chessboard) or ad-hoc calibration patterns [17, 18, 37, 58, 59, 68, 76, 80, 100] which are placed on the road plane. The main drawback of this technique is that it requires stopping traffic on the road or at the lane level temporarily. Only a few approaches tackle the calibration of both intrinsic and extrinsic parameters in an automatic fashion. The most common approach is to calculate two or three orthogonal vanishing points from two or three sets of parallel lines perpendicular to each other [133]. As described in [134], these vanishing points can be extracted from both static and moving (vehicles in this case) elements present on the scene [23, 33, 85]. In [77] the license plate tracked over time is used as a calibration pattern to hard calibrate the system. Most sophisticated approaches combine fine-grained classification to determine the specific vehicle make and model [135] (and therefore, their real dimensions) with vehicle keypoint/pose detection [136] to perform either soft [90] or hard [101] system calibration. 

Regardless of the method used in the calibration, either to calculate the translation vector, or the complete homography, it is necessary to know a priori the dimensions of some feature/object in the real world. In fact, computing the so-called \textbf{scale factor (m/px)} to transform from pixel to real-world coordinates is one of the key problems to be solved when dealing with distance and speed estimation using monocular systems. The most common features are the road/lane width or the length of a section/region manually measured [19, 24, 38, 47, 54, 64, 66, 67, 72-75, 79, 82, 84, 87, 88, 89, 98, 101, 107, 113, 128, 129], the size of a previously known object such as vehicles [27, 62, 78, 85, 90, 103, 105, 106, 110, 117] or license plates [60, 66, 68, 70, 80, 94, 133], and the length and frequency of the lane markings [14, 20, 22, 23, 26, 61, 65, 78, 108, 119, 121].

Finally, we have to consider stereo-based approaches, which are hard calibrated in the lab or on-site using standard stereo calibration techniques and patterns [60, 74, 92, 100, 112]. In addition to the intrinsic matrix $K$ per each camera, stereo systems requires the computation of the fundamental matrix which includes the extrinsic transformation from one camera to the other one. Thanks to the ability of stereo systems to obtain 3D real-world measurements from 2D pixels in both images (after solving the correspondence problem) there is no need to compute the extrinsic transformation between the road plane and the cameras, but, in any case, it can be computed manually [137] or automatically computed [138].

\section{Vehicle detection and tracking}\label{sec:t2}

This section provides a brief overview of the different vision-based vehicle detection and tracking approaches used to measure vehicle speed. However, the techniques applied to address vehicle detection and tracking as the first stages of vehicle speed detection systems are practically the same as those used for traffic surveillance applications. Therefore, we refer to [4] for a more in-depth and specific analysis of the different approaches. 

\subsection{Vehicle detection}
\subsubsection{Static background}
With the exception of drone-based systems, most approaches are based on static cameras, making the vehicle detection task much easier thanks to the presence of a static background. We can even find a considerable number of works applying the simplest frame-by-frame approach followed by a thresholding method to perform image segmentation [14, 32, 34, 36, 37, 40, 42, 46, 54, 73, 103]. The most common approach to perform vehicle detection is based on background subtraction [12, 13, 15, 18, 20-28, 30, 36, 37, 43, 44, 47, 48, 50, 51, 53, 55, 61-63, 65, 71, 72, 74, 75, 81, 83, 84, 86, 92, 93, 95, 102, 108, 110-113, 117 118, 121], followed by some morphological operations and a blob analysis method. Different methodologies are used to perform background subtraction, including gray- and color-based approaches, Gaussian Mixture Models [23, 61, 84, 102], adaptive background modeling [51, 111], etc. In some cases, frame-by-frame or background subtraction is combined with some feature detector (e.g., edges, corners, or KLT features) [40, 48, 50, 71, 110]. 

\subsubsection{Feature-based}
Some approaches are based on the detection of different type of features grouped in regions within the vehicle area, for example, after background subtraction [38, 39, 57, 99, 110, 111, 116], as well as simple methods that rely on edges [49, 87, 95], gray-level features [45, 117, 123], binary features or patterns [71, 87, 96], corners [79], SIFT/SURF features [62, 112] or KLT features [58, 59, 69, 91]. 

\subsubsection{License plate detection}
Although the appearance of the license plates may vary a little depending on the country and the type of vehicle, in general it is a homogeneous and distinctive element of any vehicle. Indeed, it is practically impossible for a detected plate not to correspond to a vehicle, and it is very unlikely that a vehicle will transit without a plate, as this is something that is criminally punishable throughout the world. This makes the license plates a very attractive target for vehicle detection systems [31, 41, 49, 60, 66, 68, 70, 76, 77, 80, 82, 94, 122, 125]. Therefore, License Plate Recognition (LPR) systems can be applied when camera resolution is large enough, i.e., from medium to low meter-to-pixel ratios. LPR systems are usually divided into at least 3 steps: detection or localization, character segmentation and character recognition. In this case, to detect a vehicle, it is only necessary to apply the first step, and there are many methods for this (we refer to some of the multiple specific surveys on this topic [139, 140]).

\subsubsection{Learning-based}
As is the case in other fields of application, learning-based approaches are recently gaining ground thanks to the availability of data and the advance in machine learning. Although the first approach based on learning was found in 2007 [19], using Boosting type methodologies (Adaboost, also used in [120], which is the only one not based on deep learning), it is not until 2017 that we see consolidated the use of this type of methodologies in all the application domains, including Faster R-CNN [85, 88, 90, 98, 101, 106, 128] an the extended version Mask R-CNN [89], SSD (Single Shot MultiBox Detector) [100, 128] and different versions of the YOLO (You Only Look Once) detector, i.e., YOLO [115], YOLOv2 [128, 129] and YOLOv3 [104, 105, 126]. In [91] transfer learning is proposed to fine-tune a vehicle detection model for their specific traffic scenario.

\subsection{Vehicle tracking}
Speed detection is a multi-frame problem. Once the vehicles or some parts/features of them are detected, a tracking method is usually applied to filter and smooth noisy detections. Tracking implicitly involves solving the data association problem (e.g., Hungarian algorithm), especially for cases with multiple objects [148]. In general terms, two main dimensions can be considered when describing the different tracking approaches. First, in terms of the type of element and/or parts of it being tracked (approaches), and second, in terms of the specific methodology used for carrying out the tracking (methodologies).

\subsubsection{Approaches} 
\begin{figure}[!t]
	\centering	
	\includegraphics[width=0.48\textwidth]{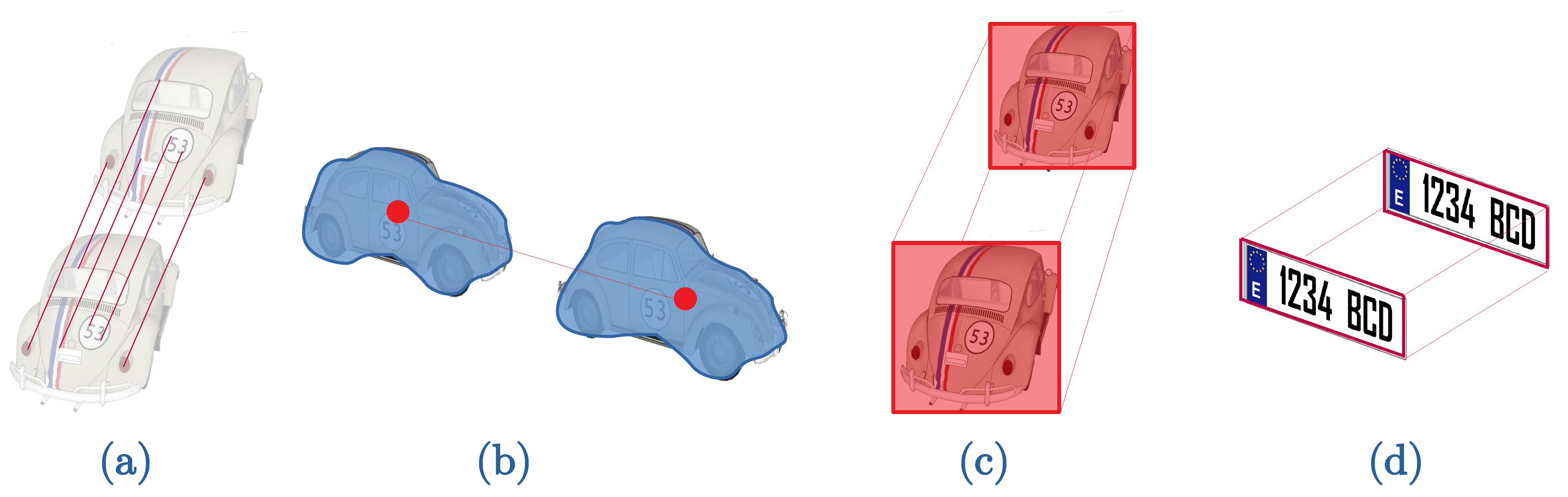}
	\caption{Vehicle tracking approaches. (a) Feature-based. (b) Blob centroid tracking. (c) Bounding box tracking. (d) License-plate-based.}
	\label{fig:track}
\end{figure}

A visual representation of the main vehicle tracking approaches is presented in Fig. \ref{fig:track}. As can be observed we can classify the different approaches into four main categories. First, \textbf{feature-based} approaches which pose the vehicle tracking problem as a feature tracking problem. In most cases, the features are either binary or gray level values within the vehicle region after applying background subtraction or a learning-based method [20, 37, 38, 45, 51, 61, 74, 85, 92, 93]. KLT features or corners are usually considered when tracking is posed as an optical flow problem, analyzing motion vectors within the vehicle region [19, 33, 53, 55, 58, 59, 69, 79, 88, 91, 97]. Other tracked features are binary patterns [71], edges [55, 79] or SIFT/SURF descriptors [62, 75, 112].

Second, we can find methods focusing on \textbf{tracking the centroid} of the region representing the vehicle, which can be considered as the contour (either convex hull or concave) [48, 65, 72, 73, 111, 121], or some model such as the convex hull of the contour [40], an ellipse [75] or the bounding box fitted to the detected blob [34, 35, 50, 56, 83, 86, 95, 103, 106, 118]. In any case, the use of the centroid as a representative state of the vehicle is very unreliable. On the one hand, most of the blob detection approaches are not very accurate and the vehicle contour may vary in the sequence due to multiple factors (shadows, overlaps, close objects, etc.). This makes the centroid rarely representing the same point of the vehicle in time. But even in those cases in which the blob detection is very precise (e.g. learning-based) the changes of perspective imply considerable variations that affect the location of the centroid. Only in very particular cases where detection is accurate and perspective changes are not relevant (for example, in drone based systems), is the use of this approach justified. 

A third group of approaches are those that use the \textbf{entire region of the vehicle} (contour- or bounding box-based) to perform the tracking. This is by far the most widely used approach [15, 16, 18, 26-30, 32, 36, 42-44, 47, 54, 56, 57, 63, 78, 81, 84, 91, 98, 102, 104, 105, 108, 109, 111, 113, 114, 120, 125, 128, 129]. Finally, a number of approaches focus on the \textbf{tracking of the license plate} [41, 49, 60, 66, 68, 70, 76, 77, 80, 82, 94, 100, 125]. Somehow this type of approaches do not calculate the speed of the vehicles but of the license plates, something completely equivalent and perfectly valid since the movement of the plate is consistent with that of the vehicle. 

\subsubsection{Methodologies}
Most of the works address the vehicle tracking problem following a \textbf{tracking by detection} methodology [15, 17, 18, 20, 25, 31, 32, 34, 38, 40, 42-45, 47, 48, 50, 51, 55, 56, 62, 71-75, 81, 83, 86, 91-93, 98, 102, 105, 106, 111, 112, 118, 120, 121, 125, 128] by means of template, regions or feature matching/correlation methods, by detecting the intersection of the vehicles over predefined regions or intrusion lines, among others. These methods solve the data association problem (associating a unique identifier to each object for each track), but do not filter the state of the tracked object. A particular case of tracking by detection methods is that of license plates, since they allow the use of Optical Character Recognition (OCR) systems as a support for solving the data association problem, making the tracking system more robust [41, 49, 60, 66, 68, 70, 76, 77, 80, 82, 94, 100, 125]. 

The use of KLT features for vehicle detection is usually followed by a KLT tracked to detect the \textbf{optical flow} [19, 53, 54, 58, 59, 69, 79, 84, 88, 91, 97, 105]. This method can only be applied when the vehicle displacement between the images is small. As we will see in the Section \ref{sec:speederror}, this is not appropriate to obtain accurate speed measurements. 

The use of Bayesian filtering is limited to the classic linear \textbf{Kalman filter} [22, 27, 61, 65, 78, 85, 104, 109] which is usually applied considering constant velocity models (less applicable for very large road sections). In [89] Simple Online and Real Time (SORT) (which is based on the Kalman filter) tracking and the extended version DeepSORT are also applied. 

\section{Vehicle distance and speed estimation}\label{sec:t3}

\subsection{Fundamentals of camera-based distance error}
\label{sec:disterror}
Cameras are sensors that generate a discretized representation of the 3D world projected into a 2D plane. Considering the solid angle between the optical center and a pixel, the area in real-world coordinates, $A_w$, increases quadratically with distance $Z$:

\begin{equation}
A_w = \frac{A_{px}Z^2}{f^2}
\end{equation}
where $f$ is the focal distance and $A_{px}$ is the area of a pixel. Even with perfect calibration of the intrinsic parameters of the camera, and with a zero pixel localization error, the uncertainty or error of distance measurements also increases quadratically with object distance for both monocular (e.g., with prior knowledge about object width [80] or flat road assumption [131]) and stereo systems [132]. Distance error for monocular systems can be formulated as:

\begin{equation}
Z_{err_m} = \frac{Z^2}{f_x \Delta X}n_m
\label{eq:3}
\end{equation}
where $f_x$ is the focal distance in $x$ axis in pixels, $\Delta X$ is the known object dimension (e.g., the width of the car [131] or the license plate [80]), and $n_m$ represents the pixel detection error. A similar expression can be formulated for stereo systems:

\begin{equation}
Z_{err_s} = \frac{Z^2}{f_x B}n_s
\label{eq:4}
\end{equation} 
where $B$ is the baseline or distance between the cameras, and $n_s$ is the disparity error. 

According to the previous expressions, we can conclude that to minimize the error in the estimation of distances of objects or features of these, it is required to:

\begin{itemize}
\item Use telescopic lenses (high focal length), since the error decreases with the focal distance. 
\item Use large stereo baselines in stereo systems, or use objects of known dimensions as large as possible in monocular systems. 
\item Avoid taking measurements of vehicles when they are far away from the camera. The closer the object, the smaller the error in distance estimation. 
\end{itemize}

The examples depicted in Fig. \ref{fig:cameras} assume these requirements from less (Fig. \ref{fig:cameras}.a) to more (Fig. \ref{fig:cameras}.c) severity.

\subsection{Relative distance estimation}

\begin{figure}[!t]
	\centering	
	\includegraphics[width=0.49\textwidth]{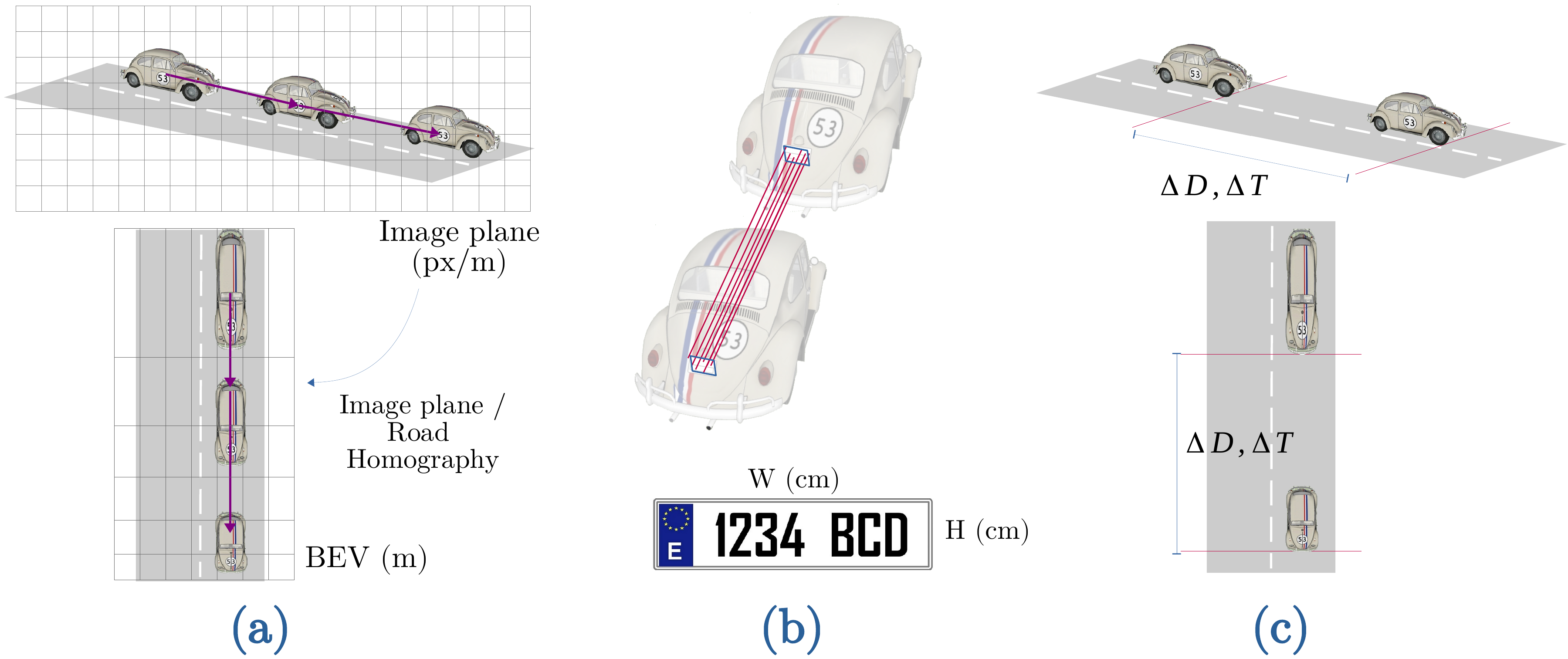}
	\caption{Distance estimation approaches: \textbf{(a)} image plane-road plane homography (px/m transformation), \textbf{(b)}  prior knowledge of the real-world size of the license plate, the vehicle or any other object; \textbf{(c)} intrusion or augmented lines, vehicle intersection with the road.}
	\label{fig:distance}
\end{figure}

\subsubsection{Stereo-based approaches}

Measuring distances in real-world coordinates is the most critical problem to be solved for accurate vehicle speed estimation. This task is straightforward when using stereo vision [42, 60, 61, 74, 80, 92, 100, 112]. For each detected vehicle, relative distances can be directly obtained using the disparity values of the pixels contained in the region of the vehicle. However, as established in subsection \ref{sec:disterror}, the accuracy of the measurements given by the stereo system is compromised by several factors such as the distance of the vehicle with respect to the stereo pair of cameras, the baseline between the cameras and the disparity error. The number of proposals using stereo cameras to deal with vehicle distance estimation is somehow limited due to the strong calibration requirements and the fact that most of the traffic cameras already installed are monocular systems. 

\subsubsection{Monocular-based approaches}
As described in subsection \ref{subsec:cal}, the previous knowledge of the dimension in real-world coordinates of some features, objects, or parts/sections of the road is a fundamental issue for the estimation of distances using monocular systems. This is usually referred to as the \emph{scale factor} for transforming pixels to real-world coordinates. Another common requirement is to consider the flat road assumption. 

\begin{figure}[!t]
	\centering	
	\includegraphics[width=0.49\textwidth]{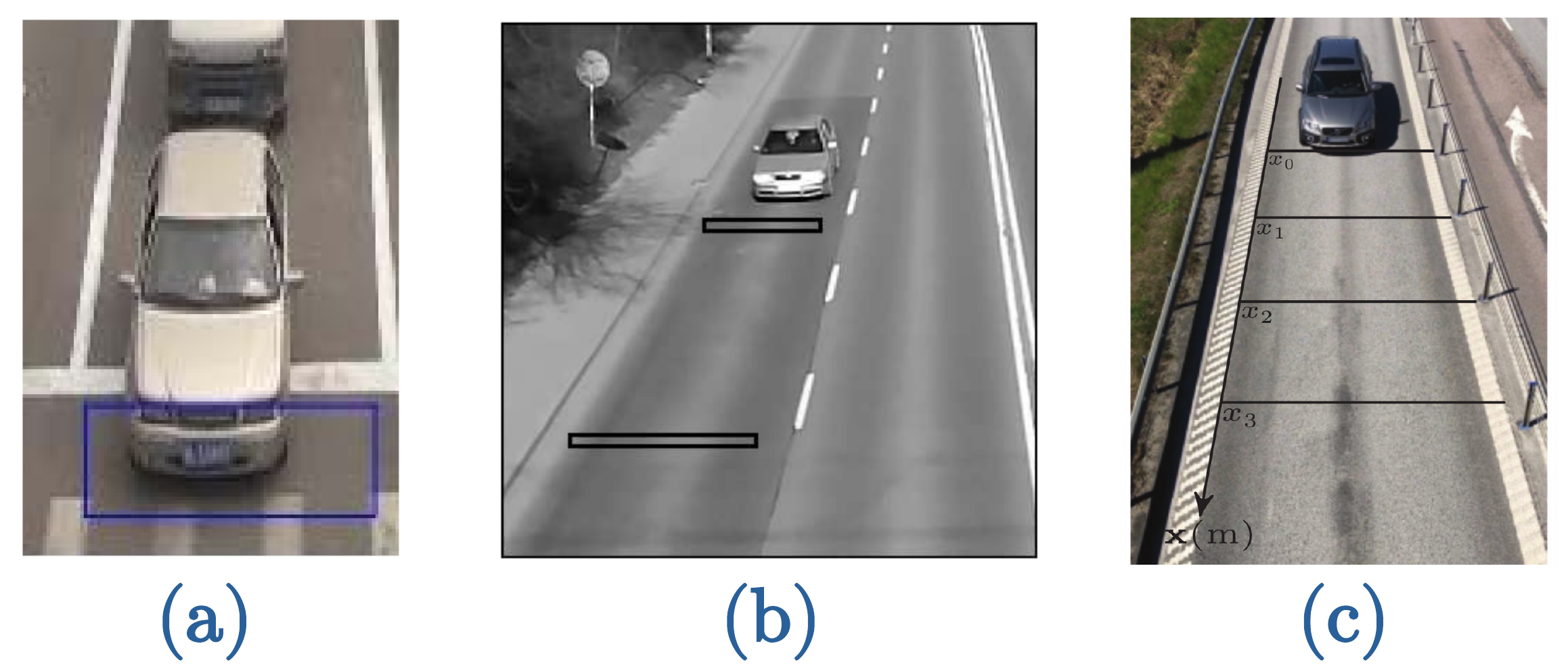}
	\caption{Distance estimation based on previous knowledge of the dimensions of a \textbf{(a)} road section [47], \textbf{(b)} two [87] and \textbf{(c)} multiple intrusion or augmented lines [107].}
	\label{fig:intru}
\end{figure}

As depicted in Fig. \ref{fig:distance} we categorize monocular-based distance estimation approaches into three groups. First, based on \textbf{intrusion or augmented lines or regions} [19, 20, 23, 24, 26, 28, 32, 38, 47, 59, 79, 86, 87, 99, 107, 124]. These approaches do not require calibration of the camera system but measure the real distance between two or multiple virtual lines on the road, or the actual size of a road region (see Fig. \ref{fig:intru}). Then, the problem of distance estimation is posed as a detection problem in which all vehicles are detected at the same distances whenever they cross the predefined virtual lines or regions. Since the virtual lines or regions are placed on the road, accurate distance estimation involves the accurate location of the contact point of some part of the vehicle. This part of the vehicle should be the same at the second location to obtain a coherent estimation of the speed. This is a complex task to be solved due to perspective constraints, spatial (depending on the camera resolution and camera-to-vehicle distance) and temporal (depending on the camera frame rate and the vehicle speed) discretization problems, shadows, etc. These problems can be mitigated by using multiple intrusion lines [20, 107].

\begin{figure}[!t]
	\centering	
	\includegraphics[width=0.49\textwidth]{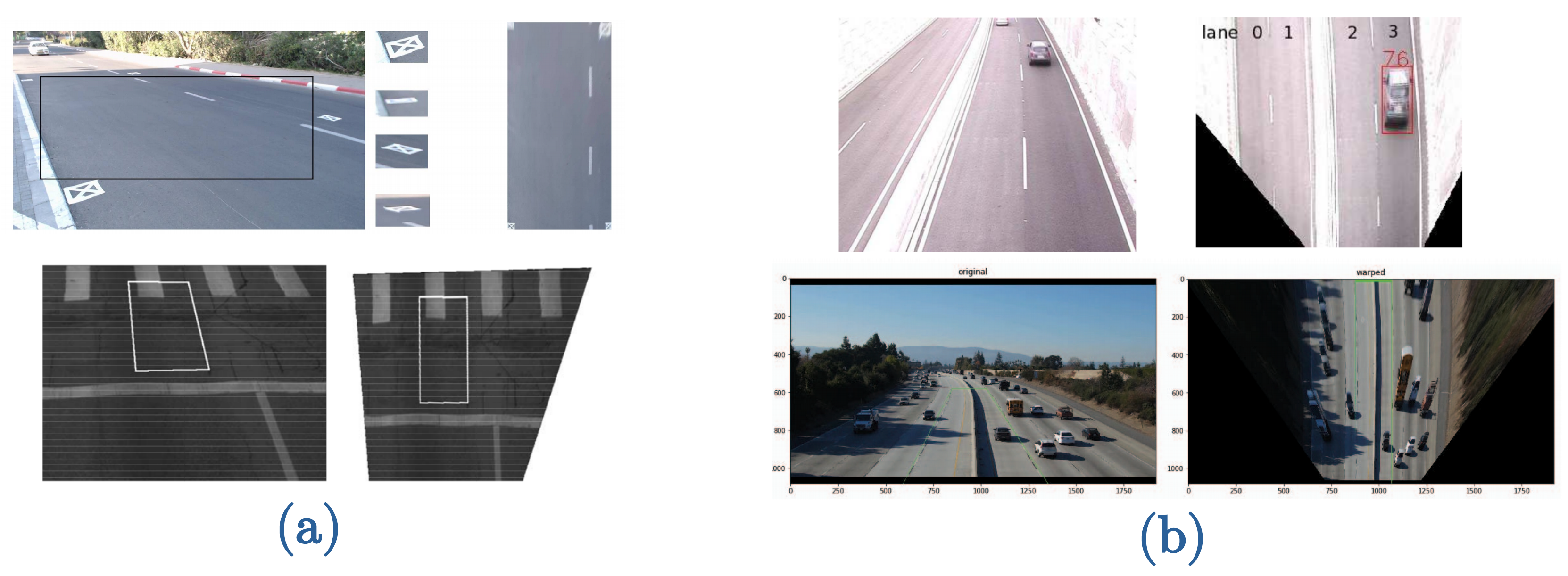}
	\caption{Distance estimation based on homography. \textbf{(a)} Calibration using previous knowledge of the real-world dimensions of a road section [66, 53]. \textbf{(b)} Examples of the BEV or image warping [22, 98].}
	\label{fig:homo}
\end{figure}

\begin{figure*}[!t]
	\centering	
	\includegraphics[width=0.95\textwidth]{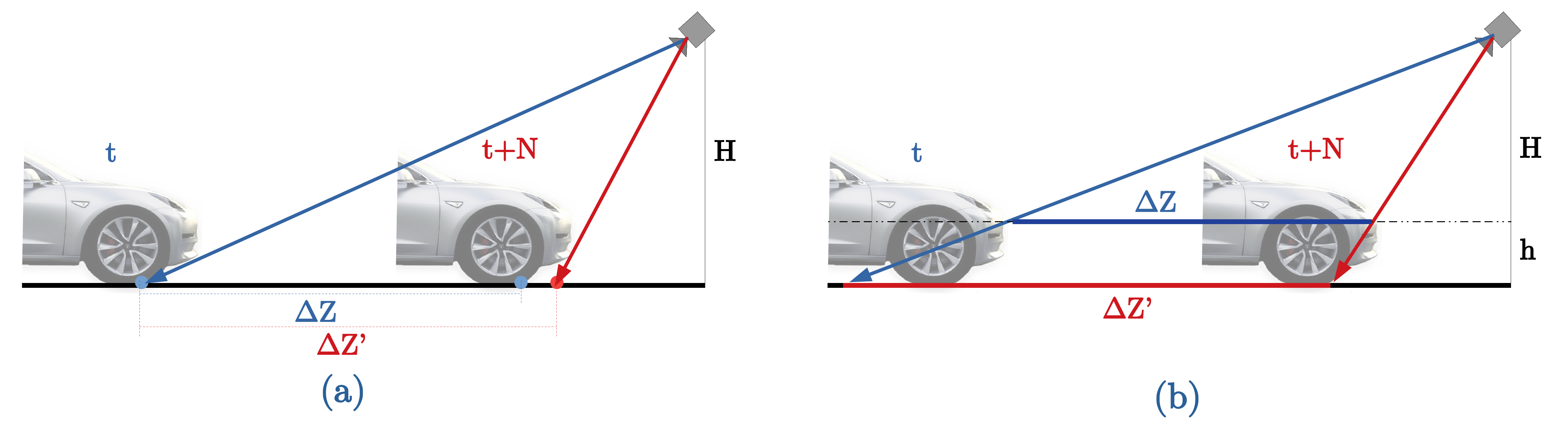}
	\caption{Illustration of distance/speed errors in homography-based approaches. \textbf{(a)} Perspective issues in which the distance measured by the camera system $\Delta Z'$ does not correspond to the same point. $\Delta Z$ denotes the real distance which cannot be obtained since the contact point is not visible in $t+1$. This problem also appears when using intrusion lines. \textbf{(b)} Wrong assumption of motion on the road plane. $\Delta Z$ denotes the real distance traveled by, for example, the license plate (which takes place at height $h$), and $\Delta Z'$ corresponds to the wrong estimation of the distance traveled by the license plate after being projected to the road plane using the homography transformation.}
	\label{fig:speedError}
\end{figure*}

Second, we have those approaches based on the computation of the \textbf{homography} as a linear transformation of a plane (the road) from the 3D projective space to a 2D projective space (the camera image plane). This way the images can be transformed into a bird's eye view (BEV), which is also known as image warping, in which pixel displacements can be directly transformed to real-world distances. This approach is the most commonly used for both fixed systems [15, 16, 18, 20, 22, 34, 37, 52, 53, 61, 66, 94, 98, 99, 110] and drone-based [25, 30, 57, 62, 64, 69, 96, 104-106, 126]. As depicted in Fig. \ref{fig:homo}, and explaining in subsection \ref{subsec:cal}, the computation of the homography involves the calibration of the intrinsic and extrinsic parameters of the camera and previous knowledge of the real dimensions of some part of the scene. The resulting warped image considers all objects to be on the road plane, which implies deformations of the shape of the vehicles. This is not particularly important, since the points used to calculate the distance generally correspond to the contact points between the vehicles and the road, although this assumption is not entirely accurate due to the perspective of the system and the wrong assumption of motion on the road plane, as depicted in Fig. \ref{fig:speedError}. The main advantage of homography-based approaches compared with the use of intrusion lines is that vehicle detection can be made at any point on the road, and therefore camera frame rate and vehicle speed do not have a negative effect om distance estimation. However, a very precise system calibration is required. 

Third, we have the approaches based on the formulation of the estimation of distances from the \textbf{knowledge of the real dimensions of the objects}, including license plates [41, 49, 60, 66, 68, 80, 125] and vehicles [42, 62, 78, 103]. On the one hand, approaches based on license plate dimensions are very sensitive to low meter-pixel ratios and pixel localization errors. On the other hand, such approaches based on vehicle dimensions of the vehicles need to apply fine-grained car model classification systems [135] to identify the specific model and use its specific dimensions.

\subsection{Speed error basics}
\label{sec:speederror}

In a simple scenario, speed is computed using two different locations $Z_1$ and $Z_2$ (which represents a linear segment $\Delta Z=|Z_2-Z_1|$) of a vehicle detected at times $t_1$ and $t_2$ ($\Delta t=|t_2-t_1|$) respectively. The average speed of that vehicle will be estimated as:

\begin{equation}
v=\frac{|Z_2-Z_1|}{|t_2-t_1|} = \frac{\Delta Z}{\Delta t}
\label{eq:speed}
\end{equation} 

Taking into account the distance estimation errors at the two locations $Z_{1err}$ and $Z_{2err}$, in the worst case we could obtain a speed of:

\begin{equation}
v'=\frac{\Delta Z + Z_{2err} + Z_{1err}}{\Delta t}
\end{equation} 
resulting in an absolute speed error of:

\begin{equation}
v_{err} = |v'-v|=\frac{|Z_{2err} + Z_{1err}|}{\Delta t} = \frac{|Z_{2err} + Z_{1err}|}{\Delta Z}v
\label{eq:verr}
\end{equation} 
being the relative speed error equal to:

\begin{equation}
\frac{v_{err}}{v} = \frac{|Z_{2err} + Z_{1err}|}{\Delta Z}
\label{eq:verrrel}
\end{equation} 

We can merge Eqs. (\ref{eq:3}) or (\ref{eq:4}) in Eqs. (\ref{eq:verr}) and (\ref{eq:verrrel}) to see that the speed error is proportional to the square of the distance from the camera to the vehicle, which clearly reinforces the requirement that speed measurements should be made with the vehicle as close to the camera as possible. In addition, Eqs. (\ref{eq:verr}) and (\ref{eq:verrrel}) reveal the importance of road segment size in the accuracy of speed estimation. That is, the quadratic penalty with the distance is compensated linearly with the length of the road section, which for segments much larger than the distance errors, results in a speed error that grows linearly with the relative distance of the vehicles. This is the main reason why average vehicle speed cameras are so accurate. These systems are based on two cameras located several kilometers apart. To meet the requirements of Eq. (\ref{eq:req}), the errors in measuring relative vehicle distances can be several meters and still, maintain a relative speed error within the range of $1-3\%$.

In the case of fixed speed cameras, the main problem of using the largest possible section of the road is that we would be considering measurements at long distances, with the mentioned penalty in the speed error. Following the procedure proposed in [80], if we differentiate Eq. (\ref{eq:verr}) with respect to the relative vehicle distance, \textbf{an optimal value for the section of the road can be obtained where the speed error is the minimum possible}. This optimal stretch of road may fall outside the limits of the image plane when using one camera. That is why a two-camera-based system is proposed in [80] with different focal lengths, similar to the one presented in [61] but pointing at closer distances. 

\begin{figure}[!t]
	\centering	
	\includegraphics[width=0.4\textwidth]{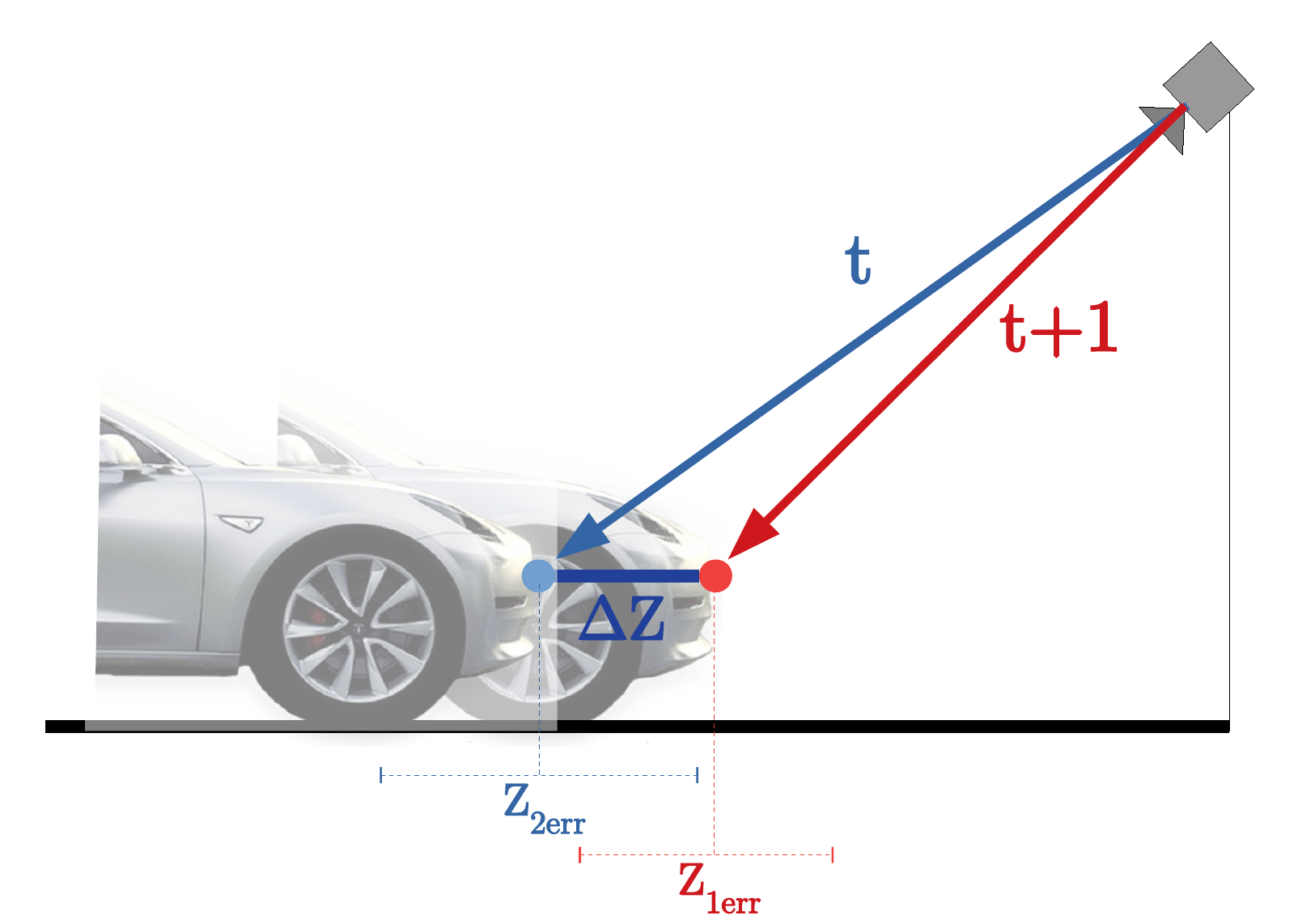}
	\caption{Illustration of the speed errors when using consecutive frames. The distance traveled by the detected and tracked features $\Delta Z$ (the vehicle) has the same order of magnitude as the distance errors $Z_{2err}$ and $Z_{1err}$, giving an inaccurate estimate of speed.}
	\label{fig:consecu}
\end{figure}

The size of the road section used to calculate vehicle speed is also directly related to the time between measurements. Therefore, when using consecutive images, even with low camera frame rates, the length of the segment $\Delta Z$ can be about the same order of magnitude as the distance errors $Z_{err}$, making it almost impossible to obtain accurate speed measurements (see Fig. \ref{fig:consecu}). 

Summarizing, we can state the following requirements to minimize the speed error:

\begin{itemize}
\item Use distance measurements of the vehicles when they are closer to the camera. 
\item Avoid using consecutive measurements with short sections. 
\item When possible, use the optimal road segment to minimize the speed error. 
\item If the optimal road section falls outside the image plane, use the largest possible road stretch.
\end{itemize}

Other sources of errors include the violation of the flat-road and constant speed assumptions, which are especially relevant when measuring speed on long stretches of road.

\subsection{Speed computation}
\subsubsection{Traffic and individual speed}
As described in the proposed taxonomy, the first possible categorization refers to the application domain. On the one hand, we have such approaches focused on measuring the average speed of all vehicles along a road segment, resulting in the \textbf{traffic speed}, or approaches where the accuracy of the measured speed of each \textbf{individual vehicle} is not so important as they are intended to be used as a traffic speed measurement system (with an application context that allows higher margins of error) [12, 17, 20, 23, 32, 34, 38, 40, 42, 51, 53, 54, 58, 61, 69, 77, 79, 80, 81, 84, 85, 94, 102, 105, 110, 113, 121]. The rest of the approaches we found are based on obtaining the precise speed of each individual vehicle. Obviously, when these approaches are based on systems that cover all lanes of the road, they can be used to obtain the average speed of the traffic. 

\subsubsection{Time/distance between measurements}
Once the relative distances of the vehicle, or part of it, have been computed, the speed calculation involves the application of Eq. \ref{eq:speed}. The first requirement is, therefore, related with associating a specific timestamp per each measurement (or image). Prior knowledge about the camera frame rate can be used. However, it is much more accurate and robust to include specific timestamps per each recorded image or measurement taken from the recording system clock (which can be globally and accurately synchronized using NTP servers [80]). 

Depending on the camera pose w.r.t. the vehicles and the road, and the speed of the vehicles, a set of $N$ measurements are available per each vehicle. The first decision to be made is which distance between measurements to use to obtain the speed values. 

Surprisingly, and contrary to the evidence described in subsection \ref{sec:speederror}, most of the works propose to use \textbf{consecutive} measurements (at frame $t$ and at frame $t+1$) to calculate the speed values for each vehicle [14, 18, 19, 22, 25, 33, 35-37, 41, 46, 48, 50, 55, 65, 67, 68, 71-76, 78, 82, 83, 85, 88, 89, 92, 96, 100, 103, 120, 129]. As illustrated in Fig.\ref{fig:consecu}, this technique tends to provide noisy values since the distance errors are of the same order of magnitude as the distance traveled by the vehicle. In the case of the use of \textbf{non-consecutive} images, we distinguish between several techniques. First, using a fixed distance or region between measurements (e.g., approaches using intrusion lines or regions) [19, 20, 23, 24, 26, 28, 32, 38, 47, 59, 86, 87, 99, 107, 124]. Second, using a predefined number of frames between measurements [15, 52, 93, 108, 111, 125] or a fixed time interval [31, 49, 111] (which are equivalent). Other approaches use the maximum possible distance between the first and last detection of the tracked vehicle [43, 44, 66, 70, 79, 118, 128]. Finally, we can identify a few techniques that make use of vehicle detection from two different cameras [80] or from different views of a drone camera [106].


\subsubsection{Measurement integration}
Once all the $N$ speed values have been computed for a vehicle, the next decision to be made is how to integrate all the measurements to compute the final vehicle speed. Note that all approaches based on virtual intrusion lines or regions [19, 20, 23, 24, 26, 28, 32, 38, 47, 59, 86, 87, 99, 107, 124] are based on a \textbf{single shot} ($N=1$) so, strictly speaking, no integration of measurements is applied. This approach is commonly referred to as \textbf{average speed detection}. 

Most approaches perform \textbf{instantaneous speed detection}, i.e., the vehicle speed is calculated and provided as an output for each pair of measurements (using Eq. \ref{eq:speed}), whether consecutive or non-consecutive [18, 25, 33, 43, 44, 55, 56, 57, 60-64, 66, 67, 70-76, 79, 82-84, 88, 93, 95-97, 100, 103, 108, 111, 112, 115-118, 120, 128]. In some cases, the instantaneous speed is filtered using different techniques such as moving average [98], polynomial filtering [129] or using a specific tracker (e.g., Kalman filter) [19, 22, 53, 59, 65, 78, 84, 89, 104].

Another common approach is to compute the \textbf{mean value} for all the available ($N$) speed measurements [12, 17, 20, 23, 27, 34, 38, 40, 42, 45, 51, 53, 54, 58, 69, 77, 80, 81, 85, 92, 94, 102, 105, 110, 121, 125]. In [80] the mean is computed only using the optimal speed values that correspond to the minimum system speed error.  




\subsubsection{Other approaches}
Although it is an uncommon approach, we found some methods that use an extended camera shutter to generate \textbf{motion blur} of the vehicles which can be analyzed to obtain the speed of the vehicle from lateral views [13, 21, 39]. 

In [119] projection histograms are obtained from the result of the frame difference applied in a predefined region. A group of key bins are finally selected to represent the vehicle motion, and all the possible speeds are tested one by one. 

Recently we can find a few \textbf{data-driven approaches} based on end-to-end deep learning methods that pose the speed detection problem as a regression problem. For example, in [113] two different convolutional neural networks are trained using animated images (top view of a two lane road segment), from a simulator and synthetic images generated by a cycle-consistent adversarial network (CycleGAN). A standard convolutional architecture is used, with an output fully connected layer of size one (corresponding to the average speed of the road). In [114] vehicle speed estimation is addressed as a video action recognition problem using 3D ConvNets. Finally, a modular neural network (MNN) architecture is presented in [127] to perform vehicle type classification and speed prediction. In all these cases, the input sequence corresponds to a stretch of road with multiple lanes and vehicles, so they are more suitable for traffic speed detection (individual vehicle speed estimation is only possible when one vehicle is present in the image sequence). 

\section{Datasets and evaluation metrics}\label{sec:data-metrics}

\subsection{Ground truth generation}
In order to establish the accuracy of the various approaches to perform vision-based vehicle speed detection,it is necessary to have an appropriate ground truth. As we summarized in subsection \ref{subec:gt}, there are a variety of methods and technologies for obtaining the actual speed of the vehicles captured by the cameras. Among these technologies, the most used is the \textbf{speedometer integrated in the vehicle} [31, 35, 38, 42, 47, 79, 83, 84, 93, 96, 104, 106, 110, 112] although this is not the most accurate approach. They are greatly affected by external factors such as the size or pressure of the tires, since the technology consists of directly measuring the rotation speed of the wheel axle, or that of the gearbox axle, and then transforming it into km/h. In addition, the measurements of these speedometers are often biased (overestimated) by the car manufacturer.

Secondly we can find GPS-based (Global Positioning Systems) technologies, which can be categorized into \textbf{standard-GPS} [22, 33, 41, 48, 100, 107], \textbf{GPS-odometer} [65, 66], and \textbf{Differential GPS} [58, 64, 80]. The main difference of these technologies lies in the method of obtaining the measurements. Both GPS and DGPS provide global position measurements with which the speed calculation can be computed, taking into account the time difference between measurements. In the case of DGPS some fixed support stations, with known global coordinates, help to correct the measurements returned by the satellites. On the other hand, the GPS odometer directly returns the speed value. Of these three systems, the most precise is the DGPS with an accuracy that can reach a few centimeters.

Other ground truth generation systems that can be found in the literature are range-based sensors such as \textbf{radar} [18 54 76 108 119] or \textbf{LiDAR} [68, 76, 94 ], including \textbf{speed guns}, [36, 51, 72, 74, 76, 113] and \textbf{light barriers} [16, 17, 114]. Radar-based sensors use radio waves to determine the range and speed of objects. Radio waves from the transmitter reflect off the vehicle and return to the receiver. The most common and best known type of radar is based on the Doppler effect. On the other hand, LiDAR-based technology is capable of measuring distances by illuminating the vehicles with laser light and measuring the reflection. Differences in laser return times and wavelengths are then used to accurately computing the range. Vehicle speed guns are hand held devices based on radar or laser technology, designed to be light and portable. They can also be mounted on a tripod to avoid hand movements. Light barriers use a beam of light in a transmitter-receiver system which would serve to detect the intrusion of a vehicle into the barrier with an accuracy of less than $10$ centimeters. The main drawback of these systems is due to occlusions between vehicles in multi-lane scenarios.

The last type of technology used to generate vehicle speed ground truth are \textbf{in-pavement sensors}, among which we can differentiate between the inductive loop, which consists of a circuit in which an electric current is induced when the metal mass of a vehicle passes by, and the piezoelectric sensors, which generate electricity when they are subjected to an external mechanical force. In both cases the speed calculation is carried out using the current generated in the electric circuit [53, 77, 82, 125, 128]. 

Finally, a considerable number of works do not carry out a comparison of their vehicle speed detection results with an appropriate ground truth, or do not provide sufficient information about it, which makes it impossible to know the real accuracy of their approaches [12-15, 19-21, 23, 25-28, 34, 37, 39, 40, 43-46, 49, 50, 52, 55, 56, 59-63, 67, 70, 71, 73, 75, 81, 86-89, 95, 99, 105, 111, 117, 118, 120, 121].

\begin{table*}[!ht]
\fwprocesstable{Overview of the vehilce speed detection datasets\label{tab2}}
{\begin{tabular*}{\textwidth}{@{\extracolsep{\fill}}lcccccc}\toprule
\textbf{Dataset} & \textbf{Application domain} & \textbf{\# Lanes} & \textbf{Meter-pixel-ratio} & \textbf{Resolution and sampling rate} & \textbf{Duration} & \textbf{Speed ground truth}\\
\midrule
\textbf{AI CITY CHALLENGE} & Traffic & 2-8 & Very high & 1920x1080 @30Hz & ~40 hours & None \\
\hspace{3mm} $[$88, 89, 91, 98$]$ \\
\textbf{BrnoCompSpeed} & Speed camera & 2 & High-medium & 1920x1080 @50Hz & 20,865 tagged vehicles & Laser-based light barrier \\
\hspace{3mm} $[$85, 123$]$ \\
\textbf{UTFPR dataset} & Speed camera & 3 & Medium-low & 1920x1080 @30Hz & 5 hours & Inductive loop detectors \\
\hspace{3mm} $[$53, 82, 122$]$ \\
\textbf{QMUL junction} & Traffic & 6 & Very high & 360x288 @25Hz & 1 hour & None \\
\hspace{3mm} $[$103$]$ \\
\botrule
\end{tabular*}}{}
\end{table*}

\subsection{Available datasets}
Although vehicle speed detection is a widely studied topic, the number of datasets available is very limited compared to other research areas and domains. In Table \ref{tab2} we present an overview of the found datasets and their main features. 

The most widely used dataset for traffic and vehicle speed detection in recent years is the \textbf{AI CITY CHALLENGE} [88, 89, 91, 98]. It is part of an annual challenge proposed by NVIDIA, and each year the number of videos and samples provided varies. Specifically, for the year 2018 it consisted of three video tracks. The first one contains 27 one-minute sequences, with a resolution of 1920x1080 pixels at 30fps. The second track includes 100 15-minute sequences with a resolution of 800x410 pixels at 30fps. The last track contains 15 videos between half an hour and an hour and a half long, at 1080p at 30 fps. All these videos were recorded in normal traffic situations such as highways or intersections with a very high meter-to-pixel ratio, which make them unsuitable for accurate vehicle speed detection. Actually, no speed ground truth is provided. 

Another dataset is the \textbf{BrnoCompSpeed} [85, 123], which contains 18 videos at a resolution of 1920x 1080 pixels at 50fps, including more than $20,000$ tagged vehicles. Speed ground truth is provided using a laser-based light barrier system. The sequences were recorded in real traffic situations with a high-medium meter-to-pixel ratio. Appropriate calibration parameters are available. 

The dataset available with the lowest meter-to-pixel ratio is the one generated by the Federal University of Technology of Paran\'a [53, 82, 122] (we refer to this dataset as \textbf{UTFPR} dataset). It contains up to 5 hours of videos recorded from one camera covering three lanes in different weather conditions, with associated ground truth speeds obtained by inductive loop detectors which are properly calibrated and approved by the Brazilian national metrology agency. The images have a 1920x1080 pixel resolution and they were captured at 30 fps. The scenario of the dataset corresponds to a urban environment where the maximum speed limit was 60 km/h. 

Other datasets not specifically devised for vehicle speed detection, are the \textbf{QMUL} junction dataset [103] which contains 1 hour of real traffic images at intersections at 25 fps with a resolution of 360x288 and the one compiled by the University of Beihang in China, which contains 70 images of real traffic on highways taken by a UAV, with a resolution of 960x540 [69]. No speed ground truth is available in both cases. 

\subsection{Evaluation metrics}
To measure the speed error produced by the systems, it is necessary the use of metrics to clearly assess their accuracy. The most commonly used metric is the \textbf{Mean Absolute Error} (MAE) either absolute or relative as a percentage [16-18, 31, 32-34, 41,48-50, 65-68, 70, 72, 74, 79, 80, 83, 85, 92, 99, 102, 104, 106-108, 110-114, 117-120, 124, 125, 129]. The MAE is usually provided with other variables such as the standard deviation, or other statistical metrics, to bound the accuracy of the speed estimation system.

Apart from the MAE, we can find other metrics [14, 15, 19, 24, 77], in which the total error of a mathematical function is calculated, including and modeling the error sources of the speed detection system, such as the pixel detection error, the length measurement error, or the camera calibration error. With these known errors, and some error propagation techniques, such as uncertainty propagation using partial derivatives, the total system error can be accurately obtained.

Finally, the use of the \textbf{Mean Square Error} (MSE) and the \textbf{Root Mean Square Error} (RMSE) have been proposed in several works [38, 40, 64, 74, 82, 88-92, 98, 100-102, 105, 113, 114, 116, 129]. The MSE/RMSE measures the "distance" between the estimates and the real value of what is being measured, using a sum of square differences between these two values. In some works this method is used to measure the relative distance error, finally measuring the speed error as the absolute difference between the real values (ground truth) and the computed ones.

\section{Discussion}\label{sec:discussion}
As illustrated in the Fig. \ref{fig:years}, there has been a significant increase in the number of proposals that deal with vehicle speed estimation using computer vision techniques. This can be explained by the combination of two factors: the increasing importance of accurate and effective vehicle speed detection for speed limit enforcement and traffic monitoring to increase road safety and improve traffic conditions, and recent advances in the computer vision field. However, after the in-depth analysis carried out in this paper, we can state that the set of computer vision approaches available to solve this particular problem is still somewhat behind than in other application contexts, leaving a long way to go. This is especially applicable in the case of speed enforcement due to the challenge of strict requirements on precision and robustness.

In what follows, the state of the art is summarized and discussed, focusing our analysis on four main points: lighting/weather conditions, benchmarking, best practices and data driven challenges.

\subsection{Lighting/weather conditions}
Except for some exceptions [48], the detection of vehicle speed has been addressed in good lighting (daytime) and weather (mostly sunny) conditions. This shows that the level of maturity of the subject is still insufficient. Vehicle speed detection should be robust to challenging weather conditions, including rainy, foggy and snowy, since speed enforcement and traffic management are of great importance in such scenarios. Weather conditions can be recognized [141], the impact of bad weather on images can be removed or mitigated [142], and vehicle detection and tracking approaches can be adapted to deal with complex weather conditions [143]. 

Dealing with nighttime scenarios is also very challenging due to poor visibility conditions, blurring, and glare from vehicle headlights. Specific nighttime vehicle detection approaches can be applied [144, 145] to deal with vehicle speed detection. In addition, the use of infrared lighting can be very helpful, especially for those approaches based on the detection and tracking of license plates, since it makes their detection and segmentation easier.

We can therefore infer that accurate vision-based vehicle speed detection in challenging lighting and weather scenarios is a solid line of future research. 

\subsection{Benchmarking}
In order to evaluate and compare the accuracy of vehicle speed estimation approaches, appropriate public datasets are needed, including, at least, different testing scenarios (with different meter-pixel ratios), a wide variety of types of vehicles, and a wide range of speeds (including over-speeding cases). More advanced features would include multiple lighting (daytime/nighttime) and weather (sunny, cloudy, rainy, foggy and snowy) conditions.  

As it has been described, and despite the high number of published papers, the number of available datasets around this topic is very limited. On the one hand, this shows a still low level of maturity of the subject. On the other hand, this can be also explained due to the high complexity and cost of the data recoding and labeling processes. In fact, apart from the conditions mentioned above, the datasets must also contain, for each vehicle, an appropriate labeling, including the location of the vehicles or any of their features or regions for all the images from the sequence, and a speed value (ground truth) obtained from some precision system (e.g., range-based sensors, DGPS, inductive loop detectors, etc.) whose measurements must be properly synchronized with the images captured by the cameras. 

Datasets must be generated following the best practices described in subsection \ref{subsec:best}. Appropriate system calibration parameters (intrinsic and extrinsic) should be also provided for each testing scenario. Other useful variables to be labeled can be the vehicle type, or even the make and model. 

Benchmarking also involves a good definition of accuracy metrics. Besides using MAE or MSE/RMSE metrics, it is also important to decide about the metric formulation, i.e., instantaneous or averaged from a set of consecutive or non-consecutive frames, or single shot from a fixed region. This formulation will strongly depend on the methodology used to obtain the ground truth values and there may be limitations in measuring the performance of certain approaches. For example, if we use light barriers in two different locations, instantaneous speed detection approaches could not be properly evaluated. 

The difficulties mentioned should not discourage researchers since the progress of vision-based vehicle speed detection depends largely on the availability of public datasets with the features described above. Future research projects related to this problem should allocate sufficient resources to create these datasets. 

\subsection{Best practices}
\label{subsec:best}
The evidence for the lack of an adequate benchmark can be illustrated by the following statement: we have not found a single work that provides an experimental comparison with multiple approaches. This is one of the reasons why we have not been able to relate the taxonomy of approaches to their performance. At this stage, we can evaluate best practices for future research, but unfortunately we cannot assess the actual performance of the different methods for vehicle speed detection.

Following the fundamentals of vision-based distance and speed estimation errors described in the subsections \ref{sec:disterror} and \ref{sec:speederror}, a set of requirements have been identified that must be met to address vehicle speed detection as accurately as possible. On the one hand, it is advisable to design a system that provides the lowest possible meter-pixel ratio by using high focal lenses, large baselines for stereo systems, high resolution cameras, and obtaining the distance measurements of the vehicles as close as possible. On the other hand, to minimize the speed error it is important to avoid the use of consecutive frames to calculate speed values, using optimal road segments when possible, or the largest available stretch of the road if the optimal one is not possible with a single camera. 

Although there has been significant progress in approaching the vehicle speed detection problem using computer vision techniques, surprisingly, these requirements are not given in most of the papers reviewed. However, we believe that future approaches should take them into account in order to develop accurate and robust vision-based vehicle speed solutions. 


\subsection{Data driven challenges}
The impact that data and learning-based approaches have had in other areas has definitely not taken place in the area of vehicle speed detection. Only recently have we found a few works that address this approach, with still preliminary results. The main current limitation is the need for much more data than is currently available. Machine learning-based approaches, and especially deep learning-based ones, require a large amount of data to be able to train complex models capable of solving and generalizing a regression problem from input video sequences.

We can anticipate that a possible solution to this problem is the use of synthetic datasets generated from simulators (e.g., autonomous driving simulators), since they allow almost unlimited samples to be obtained from multiple scenarios, with different camera settings (intrinsic and extrinsic), with absolute control of all the simulation variables involved (types of vehicles, speeds, etc.), including different weather and lighting conditions, and without the need for manual labeling. The main challenge will be to minimize the well known \emph{sim-to-real} gap, i.e., the applicability of the models trained with synthetic data to real world scenarios. 

Another limitation is based on the format used to provide input stimuli to the models. Currently, the few data-driven proposals found use sequences in which a single vehicle appears with a static background. The question is how to provide input data for multiple vehicles in multiple lanes. As it happens when applying video action recognition approaches for the classification of vehicle lane changes [146], it will be necessary to incorporate some type of process to generate regions of interests (ROIs) with sufficient spatial and temporal information to be able to learn to distinguish different types of speeds for different types of vehicles and scenarios. This problem can also be mitigated by using a system configuration with a very low meter-pixel ratio focusing on a single road lane.

It is expected that the number of learning-based proposals dealing with vehicle speed detection will grow in the coming years. 

\section{Concluding Remarks}\label{sec:conclusions}
In this paper, a comprehensive analysis of the vision-based vehicle speed estimation problem is presented. We review the available literature in the most important application domains: traffic monitoring and control (traffic cameras) and speed limit enforcement (speed cameras). A new taxonomy is proposed that is based on the main components of a vision-based speed detection system, including camera settings and calibration, vehicle detection and tracking, distance and speed estimation, application domains, and ground truth generation. We provide a detailed review of the most relevant work dealing with vehicle speed estimation using computer vision. Finally, we discuss the current limitations of the state of the art, and outline the most important and necessary future steps for the research community dealing with this challenging problem.  

The most important application domains are speed limit enforcement, traffic monitoring and control, and, more recently, autonomous vehicles.









\section{Acknowledgments}\label{ack}
This work has been mainly funded by research grant CLM18-PIC-051 (Community Region of Castilla la Mancha, Spain), and partially funded by research grants S2018/EMT-4362 SEGVAUTO 4.0-CM (Community Region of Madrid) and DPI2017-90035-R (Spanish Ministry of Science and Innovation).

\section*{References}\label{sec13}
\begin{enumerate}
\item[{[1]}] OIML., 'Radar equipment for the measurement of the speed of vehicles. International Recommendation' (OIML R91, 1990).\vspace*{6pt}

\item[{[2]}] de Jong, G., Daly, A., Pieters, M. et al.: 'Uncertainty in traffic forecasts: literature review and new results for The Netherlands', Transportation, 2007, 34, pp. 375--395\vspace*{6pt}

\item[{[3]}] Gates, T. J., Schrock, S. D. and Bonneson, J. A.: 'Comparison of portable speed measurement devices', Transp. Res. Rec., 2004, 1870, pp. 139–146\vspace*{6pt}

\item[{[4]}] Buch, N., Velastin, S.A., Orwell, J.: 'A review of computer vision techniques for the analysis of urban traffic', IEEE Trans. Intell. Transp. Syst., 2011, 12, (3), pp. 920–939\vspace*{6pt}

\item[{[5]}] WHO., 'Speed management. A road safety manual for decision-makers and practitioners.' (Global Road Safety Partnership, 2008).\vspace*{6pt}

\item[{[6]}] ERSO, EC., 'Speed Enforcement 2018.' (European Road Safety Observatory. 2018).\vspace*{6pt}

\item[{[7]}] Wilson, C., Willis, C., Hendrikz, J. K. et al.: 'Speed cameras for the prevention of road traffic injuries and deaths', Cochrane Database of Systematic Reviews, 2010, vol. 11, no. CD004607.\vspace*{6pt}

\item[{[8]}] Elvik, R.: 'Developing an accident modification function for speed enforcement', Safety Science, 2011, 49, pp. 920--925\vspace*{6pt}

\item[{[9]}]
Sotelo, M. A., Fern\'{a}ndez, D., Naranjo, J. E. et al.: 'Vision-based adaptive cruise control for intelligent road vehicles'. Proc. IEEE Int. Conf. Intelligent Robots and Systems (IROS) 2004, pp. 64--69\vspace*{6pt}

\item[{[10]}] Milan\'{e}s, V., Llorca, D.F., Villagr\'{a}, J. et al.: 'Vision-based active safety system for automatic stopping', Expert Systems with Applications, 2012, 39, (12), pp. 11234--11242\vspace*{6pt}

\item[{[11]}] Milan\'{e}s, V., Llorca, D.F., Villagr\'{a}, J. et al.: 'Intelligent automatic overtaking system using vision for vehicle detection', Expert Systems with Applications, 2012, 39, (3), pp. 3362--3373\vspace*{6pt}

\item[{[12]}] Schoepflin, T. N., Dailey, D. J.: 'Dynamic Camera Calibration of Roadside Traffic Management Cameras for Vehicle Speed Estimation', IEEE Trans. Int. Transp. Syst., 2003, 4, (2), pp. 90--98\vspace*{6pt}

\item[{[13]}] Lin, H-Y., Li, K-J.: 'Motion blur removal and its application to vehicle speed detection', Proc. Int. Conf.  Im. Proc. (ICIP), 2004.\vspace*{6pt}

\item[{[14]}] Cathey, F. W, Dailey, D. J.: 'A Novel Technique to Dynamically Measure Vehicle Speed using Uncalibrated Roadway Cameras'. Proc. IEEE Int. Veh. Symp. (IVS), 2005\vspace*{6pt}

\item[{[15]}] Grammatikopoulos, L., Karras, G., Petsa, E.: 'Automatic estimation of vehicle speed from uncalibrated video sequences'. Proc. Int. Symp. Modern Tech., Edu. Prof. Prac. Geo. Rel. Fields, 2005\vspace*{6pt}

\item[{[16]}] Litzenberger, M.Kohn, B., Belbachir, A. N. et al.: 'Estimation of Vehicle Speed Based on Asynchronous Data from a Silicon Retina Optical Sensor'. Proc. IEEE Intell. Transp. Sys. Conf (ITSC), 2006\vspace*{6pt}

\item[{[17]}] Bauer, D. Belbachir, A. N., Donath, N. et al.: 'Embedded Vehicle Speed Estimation System Using an Asynchronous Temporal Contrast Vision Sensor', EURASIP Journal on Embedded Syst., 2007, 82174, pp. 1--12\vspace*{6pt}

\item[{[18]}] He, X. C., Yung, N. H. C.: 'A Novel Algorithm for Estimating Vehicle Speed from Two Consecutive Images'. Proc. IEEE Workshop on App. of Comp. Vis. (WACV07), 2007\vspace*{6pt}

\item[{[19]}] Alefs, B., Schreiber, D.: 'Accurate Speed Measurement from Vehicle Trajectories using AdaBoost Detection and Robust Template Tracking'. Proc. IEEE Int. Transp. Syst. Conf. (ITSC)), 2007\vspace*{6pt}

\item[{[20]}] Zhiwei, H., Yuanyuan, L., Xueyi, Y.: 'Models of Vehicle Speeds Measurement with a Single Camera'. Proc. Int. Conf. On Comp. Intell. and Security Workshops, 2007\vspace*{6pt}

\item[{[21]}] Lin, H-Y., K-J., Chang, C-H.: 'Vehicle speed detection from a single motion blurred image', Image and Vision Comp., 2008, 26, pp. 1327--1337\vspace*{6pt}

\item[{[22]}] Maduro, C., Batista, K., Peixoto, P. et al.: 'Estimation of vehicle velocity and traffic intensity using rectified images'. Proc. Int. Conf. on Image Processing, 2008\vspace*{6pt}

\item[{[23]}] Peng, F., Liu, C., Ding, X.: 'Camera calibration and near-view vehicle speed estimation'. Proc. SPIE Image Proc.: Mach. Vis. App., 2008, 6813\vspace*{6pt}

\item[{[24]}] Li, Y., Yin, L. Jia, Y. et al.: 'Vehicle Speed Measurement Based on Video Images'. Proc. Int. Conf. Inn. Comp. Inf. and Cont., 2008\vspace*{6pt}

\item[{[25]}] Yamazaki, F., Liu, W., Vu, T. T.: 'Vehicle Extraction and Speed Detection from Digital Aerial Images'. Proc. of IEEE Int. Geo. and Rem. Sen. Symp., 2008\vspace*{6pt}

\item[{[26]}] Pornpanomchai, C., Kongkittisan, K. : 'Vehicle speed detection system', Proc. of IEEE Int. Conf. on Sig. and Image Proc. App., 2009\vspace*{6pt}

\item[{[27]}] Wu, J., Liu, Z., Li, J. et al.: 'An algorithm for automatic vehicle speed detection using video camera', Proc. of Int. Conf. on Comp. Sci. \& Edu., 2009\vspace*{6pt}

\item[{[28]}] Jing-zhong, W., Xiaoqing, X.: 'A real-time detection of vehicle's speed based on vision principle and differential detection', Proc. IEEE Int. Conf. Serv. Op., Log. Inf. (INFORMS), 2009\vspace*{6pt}

\item[{[29]}] Mehrubeoglu, M., McLauchlan, L.: 'Determination of vehicle speed in traffic video', Proc. of SPIE Real-Time Im. and Vid. Proc., 2009, 7244\vspace*{6pt}

\item[{[30]}] Liu, W., Yamazaki, F.: 'Speed detection of moving vehicles from one scene of QuickBird images', Proc. of Joint Urb. Rem. Sen. Event, 2009\vspace*{6pt}

\item[{[31]}] Tian, J., Chen, J., Wang, M.: 'Vehicle Speed Measurement Based on Images Energy', Proc. Asia-Pacific Conf. Inf. Proc., 2009\vspace*{6pt}

\item[{[32]}] Celik, T., Kusetogullari, H.: 'Solar-Powered Automated Road Surveillance System for Speed Violation Detection', IEEE Trans. on Ind. Elec., 2010, 57, (9), pp. 3216--3227\vspace*{6pt}

\item[{[33]}] Dogan, S., Temiz, M. S., Kulur, S.: 'Real Time Speed Estimation of Moving Vehicles from Side View Images from an Uncalibrated Video Camera', Sensors, 2010, 10, pp. 4805--4824\vspace*{6pt}

\item[{[34]}] Rahim, H. A. Sheikh, U. U., Ahmad, R. B. et al.: 'Vehicle Velocity Estimation for Traffic Surveillance System', Int. J. Comp., Elec., Aut., Cont. Inf. Eng., 2010, 4, (9), pp. 1465--1468\vspace*{6pt}

\item[{[35]}] Rad, A. G., Dehghani, A., Karim, M. R.: 'Vehicle speed detection in video image sequences using CVS method', Int. Journal of the Phy. Sci., 2010, 5, (17), pp. 2555--2563\vspace*{6pt}

\item[{[36]}] Rahim, H. A., Sheikh, U. U., Ahmad, R. B.: 'Vehicle speed detection using frame differencing for smart surveillance system', Proc. Int. Conf. Inf. Sci., Sig. Proc. App. (ISSPA), 2010\vspace*{6pt}

\item[{[37]}] Tan, H., Zhang, J., Feng, J. et al.: 'Vehicle Speed Measurement for Accident Scene Investigation', Proc. of IEEE Int. Conf. on E-Busi. Eng., 2010\vspace*{6pt}

\item[{[38]}] Yan, Y., Yancong, S., Zengqiang, M.: 'Research on vehicle speed measurement by video image based on Tsai's two stage method', Proc. of Int. Conf. on Comp. Sci. \& Educ., 2010\vspace*{6pt}

\item[{[39]}] Mohammadi, J., Akbari, R., Haghighat, M. K. B.: 'Vehicle speed estimation based on the image motion blur using RADON transform', Proc. of Int. Conf. on Sig. Proc. Sys., 2010\vspace*{6pt}

\item[{[40]}] Madasu, V. K., Hanmandlu, M.: 'Estimation of vehicle speed by motion tracking on image sequences', Proc. of IEEE Intell. Veh. Sym. (IVS), 2010\vspace*{6pt}

\item[{[41]}] Czajewski, W., Iwanowski, M.: 'Vision-Based Vehicle Speed Measurement Method', Proc. of Int. Conf. on Comp. Vis. and Graph. (ICCVG), 2010, pp. 308--315\vspace*{6pt}

\item[{[42]}] Nguyen, T. T., Pham, X. D., Song, J. H. et al.: 'Compensating Background for Noise due to Camera Vibration in Uncalibrated-Camera-Based Vehicle Speed Measurement System', IEEE Trans. on Veh. Tech., 2011, 60, (1), pp. 30--43\vspace*{6pt}

\item[{[43]}] Ibrahim, O., El-Gendy, H., ElShafee, A. M.: 'Towards Speed Detection Camera System for a RADAR Alternative'. Proc. Int. Conf. on ITS Communications, 2011\vspace*{6pt}

\item[{[44]}] Ibrahim, O., El-Gendy, H., ElShafee, A. M.: 'Speed Detection Camera System using Image Processing Techniques on Video Streams' Int. J. Comp. Elect. Eng., 2011, 3, (6), pp. 711--778\vspace*{6pt}

\item[{[45]}] Liang, W., Junfang, S.: 'The Speed Detection Algorithm Based on Video Sequences', Proc. on Int. Conf. on Comp. Sci. and Serv. Sys., 2012\vspace*{6pt}

\item[{[46]}] Ranjit, S. S. S., Anas, S. A., Subramaniam, S. K.: 'Real-Time Vehicle Speed Detection Algorithm using Motion Vector Technique', Proc. on Int. Conf. on Adv. in Elec. \& Electr., 2012\vspace*{6pt}

\item[{[47]}] Shen, I.Z, Zhou, S., Miao, C. et al.: 'Vehicle Speed Detection Based on Video at Urban Intersection', Research Journal of Applied Sciences, Eng. and Tech., 2013, 5, (17), pp. 4336--4342\vspace*{6pt}

\item[{[48]}] Sina, I., Wibisono, A., Nurhadiyatna, A. et al.: 'Vehicle counting and speed measurement using headlight detection', Proc. of Int. Conf. on Adv. Comp. Sci. and Inf. Sys. (ICACSIS), 2013\vspace*{6pt}

\item[{[49]}] Garg M., Goel S.: 'Real-time license plate recognition and speed estimation from video sequences', TSI Trans. Electrical Electron. Eng.,2013, 1, (5), pp. 1--4\vspace*{6pt}

\item[{[50]}] Dehghani, A., Pourmohammad, A.: 'Single camera vehicles speed measurement', Proc. of Iranian Conf. on Mach. Vis. and Imag. Proc. (MVIP), 2013\vspace*{6pt}

\item[{[51]}] Li, S., Yu, H., Zhang, J. et al.: 'Video-based traffic data collection system for multiple vehicle types', IET Intell. Transp. Sys., 2013, 8, (2), pp. 164--174\vspace*{6pt}

\item[{[52]}] Shukla, D., Patel, E.: 'Speed Determination of Moving Vehicles using Lucas-Kanade Algorithm', Int. Journal of Comp. App. Tech. and Res., 2013, 2, (1), pp. 32--36\vspace*{6pt}

\item[{[53]}] Luvizon, D. C., Nassu, B. T., Minneto, R.: 'Vehicle speed estimation by license plate detection and tracking' Proc. IEEE Int. Conf. On Acoustics, Speech and Sig. Proc. (ICASSP), 2014\vspace*{6pt}

\item[{[54]}] Lan, J., Li, J., Hu, G. et al.: 'Vehicle speed measurement based on gray constraint optical flow algorithm', Optik, 2014, 125, pp. 289--295\vspace*{6pt}

\item[{[55]}] Kumar, K. V. K., Chandrakant, P., Kumar, S. et al.: 'Vehicle Speed Detection Using Corner Detection', Proc. of Int. Conf. on Sig. and Image Proc., 2014\vspace*{6pt}

\item[{[56]}] Gupta, P., Purohit, G. N., Rathore, M.: 'Estimating Speed of Vehicle using Centroid Method in MATLAB', Int. Journal of Comp. App., 2014, 102, (14), pp. 1--8\vspace*{6pt}

\item[{[57]}] Xin, Z., Chang, Y., Li, L. et al.: 'Algorithm of Vehicle Speed Detection in Unmanned Aerial Vehicle Videos', Proc. of Transp. Res. Board 93rd Annual Meeting, 2014\vspace*{6pt}

\item[{[58]}] Qimin, X., Xu, L., Mingming, W. et al.: 'A methodology of vehicle speed estimation based on optical flow', Proc. of IEEE Int. Conf. on Serv. Op. and Log., and Inf., 2014\vspace*{6pt}

\item[{[59]}] Bandara, Y. M. P. C. Y., Niroshika, U. A. A., Abeygunawardhane, T. U.: 'Frame feature tracking for speed estimation', Proc. of Int. Conf. on Adv. in ICT for Emerging Regions (ICTer), 2014\vspace*{6pt}

\item[{[60]}] Lee, I. Ahn, J-p., Lee, E-J. et al.: 'The Vehicle Speed Detection Based on Three-Dimensional Information Using Two Cameras', Proc. of Ubi. Inf. Tech. and App., 2014, pp. 229--236\vspace*{6pt}

\item[{[61]}] Lin, L., Ramesh, B., Xiang, C.: 'Biologically Inspired Composite Vision System for Multiple Depth-of-field Vehicle Tracking and Speed Detection', Proc. of Asian Conf. on Comp. Vis. (ACCV), 2014, pp. 473--486\vspace*{6pt}

\item[{[62]}] Moranduzzo, T., Melgani, F.: 'Car speed estimation method for UAV images', Proc. of IEEE Geo. and Rem. Sensing Symp., 2014\vspace*{6pt}

\item[{[63]}] Khan, A., Ansari, I.,  Sarker, M. S. Z.: 'Speed Estimation of Vehicle in Intelligent Traffic Surveillance System Using Video Image Processing', Int. Journal of Sci. \& Eng. Res., 2014, 5, (12)\vspace*{6pt}

\item[{[64]}] Salvo, G., Scordo, A., Caruso, L.: 'Comparison between vehicle speed profiles acquired by differential GPS and UAV', Meeting of the Euro Work. Group On Transp., 2014\vspace*{6pt}

\item[{[65]}] Rao, Y. G. A., Kumar, N. S., Amaresh, S. H. et al.: 'Real-time speed estimation of vehicles from uncalibrated view-independent traffic cameras', Proc. of IEEE Region 10 Conf. TENCON, 2015\vspace*{6pt}

\item[{[66]}] Ginzburg, C., Raphael, A., Weinshall, D.: 'A Cheap System for Vehicle Speed Detection', arXiv:1501.06751, 2015\vspace*{6pt}

\item[{[67]}] Do, V., Nghiem, L., Pham Thi N. et al.: 'A simple camera calibration method for vehicle velocity estimation', Proc. Int. Conf. Elec. Eng./Electr., Comp., Telecomm. Inf. Tech. (ECTI-CON), 2015\vspace*{6pt}

\item[{[68]}] Wu, W., Kozitsky, V., Hoover, M. E. et al.: 'Vehicle speed estimation using a monocular camera', Proc. SPIE 9407, Video Surv. Transp. Imag. App., 2015\vspace*{6pt}

\item[{[69]}] Ke, R., Kim, S., Li, Z. et al.: 'Motion-vector clustering for traffic speed detection from UAV video', Proc. IEEE Int. Smart Cities Conf. (ISC2), 2015\vspace*{6pt}

\item[{[70]}] Makarov, A., Lukic, V., Choubey, B.: 'Real-Time Vehicle Speed Estimation Based on License Plate Tracking in Monocular Video Sequences', Sensors \& Transducers Journal, 2016, 197, (2), pp. 78--86\vspace*{6pt}

\item[{[71]}] Jeyabharathi, D., Dejey, D.: 'Vehicle Tracking and Speed Measurement system (VTSM) based on novel feature descriptor: Diagonal Hexadecimal Pattern (DHP)', Journal of Vis. Comm. and Image Representation, 2016, 40, pp. 816--830\vspace*{6pt}

\item[{[72]}] Sundoro, H. S.,  Harjoko, A.: 'Vehicle counting and vehicle speed measurement based on video processing', Journal of Theoretical and Applied Inf. Tech. , 2016, 84, (2), pp. 233--241\vspace*{6pt}

\item[{[73]}] Wang, J.: 'Research of vehicle speed detection algorithm in video surveillance', Proc. of Int. Conf. on Audio, Language and Image Proc. (ICALIP), 2016, pp. 349--352\vspace*{6pt}

\item[{[74]}] Jalalat, M., Nejati, M., Majidi, A.: 'Vehicle detection and speed estimation using cascade classifier and sub-pixel stereo matching', Proc. of Int. Conf. of Signal Proc. and Intell. Sys. (ICSPIS), 2016\vspace*{6pt}

\item[{[75]}] Yabo, A., Arroyo, S., Safar, F. et al.: 'Vehicle classification and speed estimation using Computer Vision techniques', Proc. of AADECA, 2016\vspace*{6pt}

\item[{[76]}] Eslami, H., Raie, A. A., Faez, K.: 'Precise Vehicle Speed Measurement Based on a Hierarchical Homographic Transform Estimation for Law Enforcement Applications', IEICE Trans. on Inf. and Sys. 2016, E99.D, (6), pp. 1635--1644\vspace*{6pt}

\item[{[77]}] Filipiak, P., Golenko, B., Dolega, C.: 'NSGA-II Based Auto-Calibration of Automatic Number Plate Recognition Camera for Vehicle Speed Measurement', Proc. European Conf.on the App. of Evolutionary Computation, 2016, pp. 803--818\vspace*{6pt}

\item[{[78]}] Han, J., Heo, O., Park, M. et al.: 'Vehicle distance estimation using a mono-camera for FCW/AEB systems', Int. Journal of Automotive Tech., 2016, 17, (3), pp. 483--491\vspace*{6pt}

\item[{[79]}] Enjat Munajat, M. D., Widyantoro, D. H., Munir, R.: 'Vehicle detection and tracking based on corner and lines adjacent detection features', Proc. of Int. Conf. on Sci. in Inf. Tech. (ICSITech), 2016, pp. 244--249\vspace*{6pt}

\item[{[80]}] Llorca, D. F., Salinas, C., Jim\'{e}nez, M. et al.: 'Two-camera based accurate vehicle speed measurement using average speed at a fixed point', Proc. of IEEE Intell. Transp. Sys. Conf. (ITSC), 2016, pp. 2533--2538\vspace*{6pt}

\item[{[81]}] Kumar, T., Kushwaha, D. S.: 'An Efficient Approach for Detection and Speed Estimation of Moving Vehicles', Proc. of Int. Conf. Of Inf. Proc. (IMCIP), 2016, pp. 726–731\vspace*{6pt}

\item[{[82]}] Luvizon, D. C., Nassu, B. T., Minetto, R.: 'A Video-Based System for Vehicle Speed Measurement in Urban Roadways', IEEE Trans. on Intell. Transp. Sys., 2017, 18, (6), pp. 1393--1404\vspace*{6pt}

\item[{[83]}] Wicaksono, D., Setiyono, B.: 'Speed Estimation On Moving Vehicle Based On Digital Image Processing ', Int. Journal of Comp. Sci. and Applied Math., 2017, 3, (1), pp. 21--26\vspace*{6pt}

\item[{[84]}] Gerat, J., Sopiak, D., Oravec, M. et al.: 'Vehicle speed detection from camera stream using image processing methods', Proc. Int. Symp. ELMAR, 2017\vspace*{6pt}

\item[{[85]}] Sochor, J., Jur\'{a}nek, R., Herout, A.: 'Traffic Surveillance Camera Calibration by 3D Model Bounding Box Alignment for Accurate Vehicle Speed Measurement', Comp. Vision and Image Understanding, 2017, 161, pp. 87--98\vspace*{6pt}

\item[{[86]}] Setiyono, B., Sulistyaningrum, D. R., Soetrisno: 'Vehicle speed detection based on gaussian mixture model using sequential of images', Journal of Physics: Conf. Series (ICoAIMS), 2017, 890\vspace*{6pt}

\item[{[87]}] Czapla, Z.: 'Vehicle Speed Estimation with the Use of Gradient-Based Image Conversion into Binary Form', Proc. Sig. Proc. Alg. Arch. Arrang. Apps. (SPA), 2017, pp. 213--216\vspace*{6pt}

\item[{[88]}] Giannakeris, P., Kaltsa, V., Avgerinakis, K. et al.: 'Speed Estimation and Abnormality Detection from Surveillance Cameras', Proc. IEEE Comp. Vis. Mach. Intell., 2018, pp. 93--99\vspace*{6pt}

\item[{[89]}] Kumar, A., Khorramshahi, P., Lin, W. et al.: 'A Semi-Automatic 2D Solution for Vehicle Speed Estimation from Monocular Videos', Proc. IEEE Comp. Vis. Mach. Intell., 2018, pp. 137--144\vspace*{6pt}

\item[{[90]}] Bhardwaj, R., Tummala, G., Ramalingam, G. et al.: 'AutoCalib: Automatic Traffic Camera Calibration at Scale', ACM Trans. Sen. Netw. 2018, 19, pp. 1--27\vspace*{6pt}

\item[{[91]}] Hua, S., Kapoor, M., Anastasiu, D. C.: 'Vehicle Tracking and Speed Estimation from Traffic Videos', Proc. IEEE Comp. Vis. Patt. Rec. (CVPR), 2018\vspace*{6pt}

\item[{[92]}] Bouziady, A. E., Thami, R. O. H., Ghogho, M. et al.: 'Vehicle speed estimation using extracted SURF features from stereo images', Proc. IEEE Int. Conf. Intell. Sys. and Comp. Vis. (ISCV), 2018\vspace*{6pt}

\item[{[93]}] Kurniawan, A., Ramadlan, A., Yuniarno, E. M.: 'Speed Monitoring for Multiple Vehicle Using Closed Circuit Television (CCTV) Camera', Proc. Int. Conf. Comp. Eng., Net. Intell. Mult. (CENIM), 2018\vspace*{6pt}

\item[{[94]}] Famouri, M., Azimifar, Z., Wong, A.: 'A Novel Motion Plane-Based Approach to Vehicle Speed Estimation', IEEE Trans. Intell. Transp. Sys., 2018, 20, (4), pp. 1237--1246\vspace*{6pt}

\item[{[95]}] Koyuncu, H., Koyuncu, B.: 'Vehicle Speed detection by using Camera and image processing software', The Int. Journal Eng. Sci. (IJES), 2018, 7, (9), pp. 64--72\vspace*{6pt}

\item[{[96]}] Long, H., Chung, Y-N., Li, J-d.: 'Automatic Vehicle Speed Estimation Method for Unmanned Aerial Vehicle Images', Journal Inf. Hid. Mult. Sig. Proc., 2018, 9, (2), pp. 442--451\vspace*{6pt}

\item[{[97]}] Patel, N. S., Raval, K. R., Agravat, S. J.: 'An Algorithm for Speed Estimation and Speed Violation Detection of a Vehicle using Computer Vision', Int. Journal Adv. Res. Comp. Sci., 2018, 9, (2), pp. 504--507\vspace*{6pt}

\item[{[98]}] Huang, T.: 'Traffic Speed Estimation from Surveillance Video Data', Proc. IEEE Comp. Vis. Patt. Rec. (CVPR), 2018, pp. 161--165\vspace*{6pt} 

\item[{[99]}] Kim, J-H., Oh, W-T. Choi, J-H. et al.: 'Reliability verification of vehicle speed estimate method in forensic videos', Foren. Sci. Int., 2018, 287, pp. 195--206\vspace*{6pt}

\item[{[100]}] Yang, L., Li, M., Song, X. et al.: 'Vehicle Speed Measurement Based on Binocular Stereovision System', IEEE Access, 2019, 7, pp. 106628--106641\vspace*{6pt}

\item[{[101]}] Bartl, V., Herout, A.: 'OptInOpt: Dual Optimization for Automatic Camera Calibration by Multi-Target Observations', Proc. of IEEE Int. Conf. on Adv. Video and Signal Based Surv. (AVSS), 2019\vspace*{6pt}

\item[{[102]}] Tourani, A., Shahbahrami, A., Akoushideh, A. et al.: 'Motion-based Vehicle Speed Measurement for Intelligent Transportation Systems', Int. Journal of Image, Graph. and Sig. Proc., 2019, 11, pp. 42--54\vspace*{6pt}

\item[{[103]}] Moazzam, M., Haque, M., Uddin, M.: 'Image-Based Vehicle Speed Estimation', Journal of Comp. and Comm., 2019, 7, pp. 1--5\vspace*{6pt}

\item[{[104]}] Liu, Y., Lian, Z., Ding, J. et al.: 'Multiple Objects Tracking Based Vehicle Speed Analysis with Gaussian Filter from Drone Video', Intell. Sci. and Big Data Eng. Vis. Data Eng. (IScIDE), 2019, pp. 362--373\vspace*{6pt}

\item[{[105]}] Li, J., Chen, S., Zhang, F. et al.: 'An Adaptive Framework for Multi-Vehicle Ground Speed Estimation in Airborne Videos', Remote Sens., 2019, 11, 1241\vspace*{6pt}

\item[{[106]}] Biswas, D., Su, H., Wang, C. et al.: 'Speed Estimation of Multiple Moving Objects from a Moving UAV Platform', ISPRS Int. J. Geo-Inf, 2019, 8, 259\vspace*{6pt}

\item[{[107]}] Javadi, S., Dahl, M., Pettersson, M. I.: 'Vehicle speed measurement model for video-based systems', Comp. \& Elec. Eng., 2019, 76, pp. 238--248\vspace*{6pt}

\item[{[108]}] Sochor, J., Juranek, R., Spanhel, J. et al.: 'Comprehensive Data Set for Automatic Single Camera Visual Speed Measurement', IEEE Trans. on Intell. Transp. Sys., 2019, 20, (5), pp. 1633--1643\vspace*{6pt}

\item[{[109]}] Wang, C., Musaev, A.: 'Preliminary Research on Vehicle Speed Detection using Traffic Cameras', Proc. IEEE Int. Conf. Big Data, 2019\vspace*{6pt}

\item[{[110]}] Gunawan, A. A. S., Tanjung, D. A., Gunawan, F. E.: 'Detection of Vehicle Position and Speed using Camera Calibration and Image Projection Methods', Procedia Comp. Sci., 2019, 157, pp. 255--265\vspace*{6pt}

\item[{[111]}] Julina, J. K. J., Sharmila, T. S., Gladwin, S. J.: 'Vehicle Speed Detection System using Motion Vector Interpolation', Proc. Global Conf. Adv. Tech. (GCAT), 2019\vspace*{6pt}

\item[{[112]}] Jiang, J., Mi, C., Wu, M. et al.: 'Study on a Real-Time Vehicle Speed Measuring Method at Highway Toll Station', Proc. Int. Conf. Sen. Inst. IoT Era (ISSI), 2019\vspace*{6pt}

\item[{[113]}] Lee, J., Roh, S., Shin, J. et al.: 'Image-Based Learning to Measure the Space Mean Speed on a Stretch of Road without the Need to Tag Images with Labels', Sensors, 2019, 19, 1227\vspace*{6pt}

\item[{[114]}] Dong, H., Wen, M., Yang, Z.: 'Vehicle Speed Estimation Based on 3D ConvNets and Non-Local Blocks', Future Internet, 2019, 11, 123\vspace*{6pt} 

\item[{[115]}] Gauttam, H. K., Mohapatra, R. K.: 'Speed Prediction of Fast Approaching Vehicle Using Moving Camera', Proc. Int. Conf. Comp. Vis. Ima. Proc. (CVIP), 2019, pp. 423--431\vspace*{6pt}

\item[{[116]}] Khan, M., Nawaz, M., Nida-Ur-Rehman, Q.: 'Multiple Moving Vehicle Speed Estimation Using Blob Analysis', Proc. World Conf. Inf. Sys. Tech. (WorldCIST), 2019, pp. 303--314\vspace*{6pt}

\item[{[117]}] Murashov, I., Stroganov, Y.: 'Method of determining vehicle speed according to video stream data', Journal of Physics: Conf. Ser., 2019, 1419, pp. 1--7\vspace*{6pt}

\item[{[118]}] Afifah, F., Nasrin, S., Mukit, A.: 'Vehicle Speed Estimation using Image Processing', Journal Adv. Res.   Appl. Mech., 2019, 48, (1), pp. 9--16\vspace*{6pt}

\item[{[119]}] Lu, S., Wang, Y., Song, H.: 'A high accurate vehicle speed estimation method', Soft Comp., 2020, 4, pp. 1283--1291\vspace*{6pt}

\item[{[120]}] Kamoji, S., Koshti, D., Dmonte, A. et al.: 'Image Processing based Vehicle Identification and Speed Measurement', Proc. of Int. Conf. on Inventive Comp. Tech. (ICICT), 2020, 4, pp. 523--527\vspace*{6pt}

\item[{[121]}] G. Cheng, Y. Guo, X. Cheng et al.: 'Real-time Detection of Vehicle Speed Based on Video Image', Proc. of Int. Conf. on Meas. Tech. and Mech. Aut., 2020, pp. 313--317\vspace*{6pt}

\item[{[122]}] Sonth, A., Settibhaktini, H., Jahagirdar, A.: 'Vehicle Speed Determination and License Plate Localization from Monocular Video Streams', Proc. of Int. Conf. on Comp. Vis. and Image Proc., 2020, pp. 267--277\vspace*{6pt}

\item[{[123]}] Mini, T. V., Vijayakumar, V.: 'Speed Estimation and Detection of Moving Vehicles Based on Probabilistic Principal Component Analysis and New Digital Image Processing Approach', Proc. of EAI Int. Conf. on Big Data Innov. for Sust. Cogn. Comp., 2020, pp. 221--230\vspace*{6pt}

\item[{[124]}] Dahl, M., Javadi, S.: 'Analytical Modeling for a Video-Based Vehicle Speed Measurement Framework', Sensors, 2020, 20, 160\vspace*{6pt}

\item[{[125]}] Vakili, E., Shoaran, M., Sarmadi, M. R.: 'Single-camera vehicle speed measurement using the geometry of the imaging system', Mult. Tools Apps., 2020, 79, pp. 19307--19327\vspace*{6pt}

\item[{[126]}] Bastos, M. E. da S., Freitas, V. Y. F., de Menezes, R. S. T.:'Vehicle Speed Detection and Safety Distance Estimation Using Aerial Images of Brazilian Highways', Proc. Semin\'{a}rio Integrado Soft. Hard., 2020\vspace*{6pt}

\item[{[127]}] Madhan, E. S., Neelakandan, S., Annamalai, R.: 'A Novel Approach for Vehicle Type Classification and Speed Prediction Using Deep Learning', Journal Comp. Theor. Nano., 2020, 17, (5), pp. 2237--2242\vspace*{6pt}

\item[{[128]}] Liu, C., Huynh, D. Q., Sun, Y.: 'A Vision-Based Pipeline for Vehicle Counting, Speed Estimation, and Classification', IEEE Trans. on Intell. Transp. Sys., 2020, Early Access, pp. 1--14\vspace*{6pt}

\item[{[129]}] Bell, D., Xiao, W., James, P.: 'Accurate Vehicle Speed Estimation from Monocular Camera Footage', Annals Phot., Rem. Sen. Spat. Inf. Sci. (ISPRS), 2020, pp. 419--426\vspace*{6pt}

\item[{[130]}] Izquierdo, R., Quintanar, A., Parra, I.: 'The PREVENTION dataset: a novel benchmark for PREdiction of VEhicles iNTentIONs.', Proc. of IEEE Intell. Transp. Sys. Conf.(ITSC), 2019, pp. 3144--3121\vspace*{6pt}

\item[{[131]}] Stein, G. P., Mano, O., Shashua, A.: 'Vision-based ACC with a single camera: bounds on range and range rate accuracy', Proc. IEEE Intell. Veh. Sym., 2003\vspace*{6pt}

\item[{[132]}] Llorca, D. F., Sotelo, M. A., Parra, I. et al.: 'Error Analysis in a Stereo Vision-Based Pedestrian Detection Sensor for Collision Avoidance Applications', Sensors, 2010, 10, pp. 3741--3758\vspace*{6pt}

\item[{[133]}] Orghidan, R., Salvi, J., Gordan, M. et al.: "Camera calibration using two or three vanishing points", Proc. Fed. Conf. Comp. Sci. Inf. Sys., 2012, pp. 123–130\vspace*{6pt}

\item[{[134]}] \'{A}lvares, S., Llorca, D. F., Sotelo, M. A.: 'Hierarchical camera auto-calibration for traffic surveillance systems', Expert Systems with Applications, 2014, 41, pp. 1532--1542\vspace*{6pt}

\item[{[135]}] Corrales, H., Llorca, D. F., Parra, I.: 'CNNs for Fine-Grained Car Model Classification', EUROCAST, LNCS, 2020, 12014, pp. 104--112\vspace*{6pt}

\item[{[136]}] Corrales, H., Hern\'{a}ndez, A., Izquierdo, R. et al.: 'Simple Baseline for Vehicle Pose Estimation: Experimental Validation', IEEE Access, 2020, 8, pp. 132539--132550\vspace*{6pt}

\item[{[137]}] Fern\'{a}ndez Llorca, D., Quintero, R., Parra, I. et al.: 'Assistive intelligent transportation systems: the need for user localization and anonymous disability identification', IEEE Int. Transp. Sys. Mag., 2017, 9, (2), pp. 25--40\vspace*{6pt}

\item[{[138]}] Fern\'{a}ndez, C, Izquierdo, R., Llorca, D. F. et al.: 'A Comparative Analysis of Decision Trees Based Classifiers for Road Detection in Urban Environments', Proc. IEEE Int. Transp. Sys. Conf., 2015, pp. 719--724\vspace*{6pt}

\item[{[139]}] Anagnostopoulos, C. E., Anagnostopoulos, I. E., Psoroulas, I. D., et al.: 'License Plate Recognition From Still Images and Video Sequences: A Survey', IEEE Trans. Intell. Transp. Sys., 2008, 9, (3), pp. 377--391\vspace*{6pt}

\item[{[140]}] Du, S., Ibrahim, M.,  Shehata M. et al.: 'Automatic License Plate Recognition (ALPR): A State-of-the-Art Review', IEEE Trans. Circ. Sys. Vid. Tech., 2013, 23, (2), pp. 311--325\vspace*{6pt}

\item[{[141]}] Yan, X., Luo,  Y., Zheng, X.: 'Weather Recognition Based on Images Captured by Vision System in Vehicle', Adv. Neural Networks - ISNN 2009, pp. 390--398\vspace*{6pt}

\item[{[142]}] Luo, Y., Xu, Y., Ji, H.: 'Removing Rain From a Single Image via Discriminative Sparse Coding', Proc. IEEE Int. Conf. Comp. Vis. (ICCV), 2015, pp. 3397--3405\vspace*{6pt}

\item[{[143]}] Rezaei, M., Terauchi, M., Klette, R.: 'Robust Vehicle Detection and Distance Estimation Under Challenging Lighting Conditions', IEEE Trans. Int. Trans. Sys., 2015, 16, (5), pp. 2723--2743\vspace*{6pt}

\item[{[144]}] Alcantarilla, P. F., Bergasa, L. M., Jim\'{e}nez, P. et al.: 'Night time vehicle detection for driving assistance lightbeam controller', Proc. IEEE Int. Veh. Symp., 2008\vspace*{6pt}

\item[{[145]}] Chen, Y-L. Wu, B-F., Huang, H-Y., et al.: 'A Real-Time Vision System for Nighttime Vehicle Detection and Traffic Surveillance', IEEE Trans. Ind. Elec., 2011, 58, (5), pp. 2030--2044\vspace*{6pt}

\item[{[146]}] Biparva, M, Llorca, D. F., Izquierdo, R. et al.: 'Video action recognition for lane-change classification and prediction of surrounding vehicles', arXiv:2101.05043, 2021\vspace*{6pt}

\item[{[147]}] Posch, C.: 'Bio-inspired vision', Topical Work. Elec. Part. Phy. 2011, IOP Publishing Ltd and SISSA, 2012\vspace*{6pt}

\item[{[148]}] Li Zhang, L., Li, Y., Nevatia, R.: 'Global data association for multi-object tracking using network flows', Proc. IEEE Comp. Vis. Patt. Rec., 2008\vspace*{6pt}

\end{enumerate}


\vfill\pagebreak

\end{document}